\documentclass[sigconf]{acmart}
\usepackage{color}
\usepackage{amsfonts}       % blackboard math symbols
\usepackage{amsmath}%, amssymb}
\usepackage{nicefrac}       % compact symbols for 1/2, etc.
\usepackage{microtype}      % microtypography
\usepackage{enumitem}

\usepackage{xspace}
\usepackage{multirow}
\usepackage{balance}
\usepackage{makecell}
\usepackage{bbding}

\usepackage{dsfont}

\newcommand{\etal}{\textit{et al.\xspace}}
\newcommand{\ie}{\textit{i.e.}, }
\newcommand{\eg}{\textit{e.g.}, }

\usepackage{pifont}% http://ctan.org/pkg/pifont
\newcommand{\xmark}{\ding{55}}%
\newcommand{\cmark}{\ding{51}}%

\AtBeginDocument{%
 }

\copyrightyear{2023}
\acmYear{2023}
\setcopyright{rightsretained}
\acmConference[MM '23]{Proceedings of the 31st ACM International Conference on Multimedia}{October 29-November 3, 2023}{Ottawa, ON, Canada}
\acmBooktitle{Proceedings of the 31st ACM International Conference on Multimedia (MM '23), October 29-November 3, 2023, Ottawa, ON, Canada}
\acmDOI{10.1145/3581783.3612000}
\acmISBN{979-8-4007-0108-5/23/10}

\usepackage{etoolbox}
\makeatletter
\gdef\@copyrightpermission{
  \begin{minipage}{0.3\columnwidth}
   \href{https://creativecommons.org/licenses/by/4.0/}
    {\includegraphics[width=0.90\textwidth]{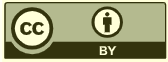}}
  \end{minipage}\hfill
   \begin{minipage}{0.7\columnwidth}
     \href{https://creativecommons.org/licenses/by/4.0/}{This work is licensed under a Creative Commons Attribution International 4.0 License.}
   \end{minipage}
  \vspace{5pt}
}
\makeatother

\begin{document}

%%
%% The "title" command has an optional parameter,
%% allowing the author to define a "short title" to be used in page headers.
\title{Prototypical Cross-domain Knowledge Transfer for Cervical Dysplasia Visual Inspection}

%%
%% The "author" command and its associated commands are used to define
%% the authors and their affiliations.
%% Of note is the shared affiliation of the first two authors, and the
%% "authornote" and "authornotemark" commands
%% used to denote shared contribution to the research.

\author{Yichen Zhang}
\affiliation{%
  \institution{National University of Singapore}
  \country{}}
\email{zhang.yichen@u.nus.edu}

\author{Yifang Yin}
\affiliation{%
  \institution{Institute for Infocomm Research, A*STAR}
  \country{}}
\email{yin_yifang@i2r.a-star.edu.sg}

\author{Ying Zhang}
\affiliation{%
  % \institution{School of Computer Science, Northwestern Polytechnical University}
  \institution{Northwestern Polytechnical University}
  \country{}}
\email{izhangying@nwpu.edu.cn}

\author{Zhenguang Liu}
\affiliation{%
  \institution{Zhejiang Gongshang University}
  \country{}}
\email{liuzhenguang2008@gmail.com}
\authornote{The corresponding author}

\author{Zheng Wang}
\affiliation{%
  \institution{Wuhan University}
  \country{}}
\email{wangzwhu@whu.edu.cn}

\author{Roger Zimmermann}
\affiliation{%
  \institution{National University of Singapore}
  \country{}}
\email{rogerz@comp.nus.edu.sg}

%%
%% By default, the full list of authors will be used in the page
%% headers. Often, this list is too long, and will overlap
%% other information printed in the page headers. This command allows
%% the author to define a more concise list
%% of authors' names for this purpose.
\renewcommand{\shortauthors}{Yichen Zhang et al.}

%%
%% The abstract is a short summary of the work to be presented in the
%% article.
\begin{abstract}
Early detection of dysplasia of the cervix is critical for cervical cancer treatment. However, automatic cervical dysplasia diagnosis via visual inspection, which is more appropriate in low-resource settings, remains a challenging problem. Though promising results have been obtained by recent deep learning models, their performance is significantly hindered by the limited scale of the available cervix datasets. 
Distinct from previous methods that learn from a single dataset, we propose to leverage cross-domain cervical images that were collected in different but related clinical studies to improve the model's performance on the targeted cervix dataset. 
To robustly learn the transferable information across datasets, we propose a novel prototype-based knowledge filtering method to estimate the transferability of cross-domain samples. We further optimize the shared feature space by aligning the cross-domain image representations simultaneously on \emph{domain level} with early alignment and \emph{class level} with supervised contrastive learning, which endows model training and knowledge transfer with stronger robustness.
The empirical results on three real-world benchmark cervical image datasets show that our proposed method outperforms the state-of-the-art cervical dysplasia visual inspection by an absolute improvement of 
4.7\% in top-1 accuracy, 7.0\% in precision, 1.4\% in recall, 4.6\% in F1 score, and 0.05 in ROC-AUC.
%3.7\% in top-1 accuracy, 9.8\% in recall, 5.3\% in F1 score, and 0.06 in ROC-AUC.
%Our code is anonymously released at \url{https://anonymous.4open.science/r/ProtoCervix-2321}.
\end{abstract}

%%
%% The code below is generated by the tool at http://dl.acm.org/ccs.cfm.
%% Please copy and paste the code instead of the example below.
%%


%%
%% Keywords. The author(s) should pick words that accurately describe
%% the work being presented. Separate the keywords with commas.

%% A "teaser" image appears between the author and affiliation
%% information and the body of the document, and typically spans the
%% page.
\begin{teaserfigure}
  \begin{center}
  \includegraphics[width=.7\textwidth]{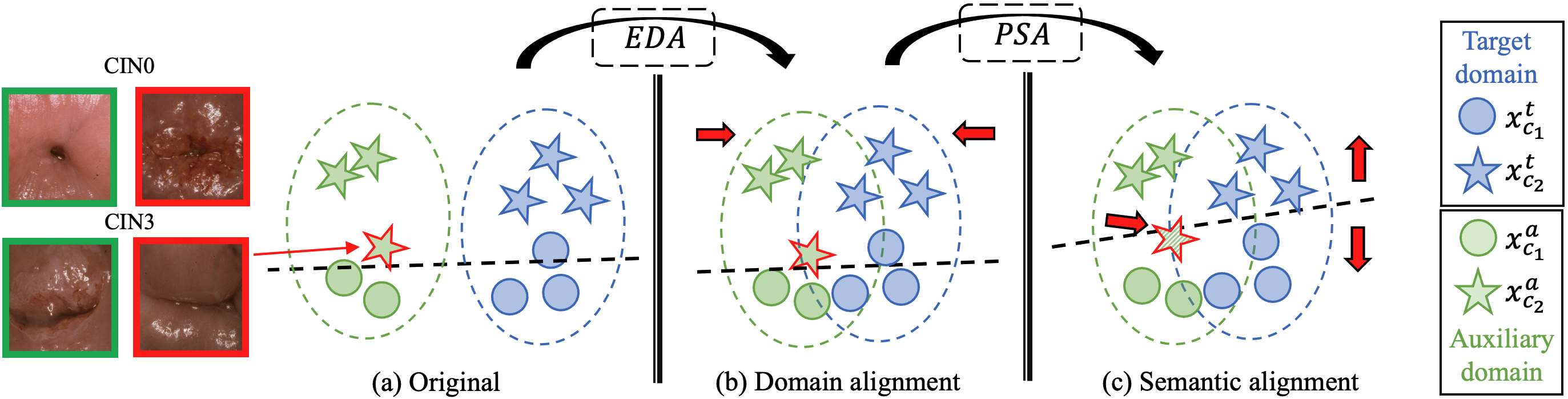}
  \end{center}
   \caption{Illustration of our proposed method. (a) In the original feature space, direct supervised learning with auxiliary samples may degrade the model's performance in the target domain. We thus propose (b) an Early Domain Alignment (EDA) module to reduce the domain gap, and (c) a Prototypical Semantic Alignment (PSA) module to identify auxiliary samples with high-uncertainty labels (\ie red border) and reduce their impact when aligning the representations at the semantic level.}
  \Description{}
  \label{fig:intro}
\end{teaserfigure}

% \received{20 February 2007}
% \received[revised]{12 March 2009}
% \received[accepted]{5 June 2009}

%%
%% This command processes the author and affiliation and title
%% information and builds the first part of the formatted document.
\maketitle
\section{Introduction}

Cervical cancer is one of the most common cancers for women~\cite{WHO-cancer}, posing serious risks to their health and spreading through direct or distant metastasis~\cite{cancer2002}. Especially in developing countries, it is the second most prevalent malignancy after breast cancer and the third dominant cause of cancer-related deaths~\cite{Li2019ASF}, despite the fact that it is one of the most successfully treatable forms of cancer if diagnosed in an early stage~\cite{Gotlieb2017ConstraintBasedVO}.
Cervical dysplasia, also known as cervical intraepithelial neoplasia (CIN), is a precancerous change indicating potential cervical cancer in an early stage. Although it can be detected via a few screening methods, %, including a Pap smear test, an HPV (human papillomavirus) test, and visual inspection. 
most of them are conducted in a laboratory setting where special infrastructure and extensively trained personnel are needed. Such constraints significantly limit their wide deployment in low-resource regions. 
To accommodate the medical needs, visual inspection of the cervix after applying 5\% acetic acid to the cervix epithelium (a method known in the medical community as VIA) has been advocated by the WHO because of its simplicity and low cost. In this paper, we focus on improving the performance of computational visual inspection to assist in faster and more accurate inspection. Note that colposcopic photographs from the VIA approach are referred to as cervical images in the rest of the paper.

Despite deep neural networks having been widely adopted in computer vision, attaining state-of-the-art performance usually requires vast quantities of labeled data. Unlike natural images, medical image acquisition, annotation, and analysis require significant efforts of human expertise~\cite{li2021systematic} and are traditionally part of localized medical studies. 
% A dearth of large task-specific datasets still stands in the way of achieving outstanding model performance.
Existing methods mostly perform transfer learning based on models pre-trained on natural images, particularly ImageNet~\cite{deng2009imagenet}, to alleviate this situation. 
While this may work well in some general instances, recent research shows that such task-agnostic transfer learning alone does not necessarily result in performance improvements for medical applications, due to the considerable visual differences between natural and medical images~\cite{raghu2019transfusion}. 
A dearth of large task-specific datasets still stands in the way of achieving outstanding model performance.

% \begin{figure}[!t]
%   \centering
%   \includegraphics[width=1.\linewidth]{fig/intro_v2.eps}
%   \caption{Illustration of our proposed cross-domain knowledge transfer method. (a) In the original feature space, direct supervised learning with auxiliary samples may degrade the model's performance in the target domain. We thus propose (b) an Adversarial Domain Alignment (ADA) module to reduce the domain gap, and (c) a Prototypical Semantic Alignment (PSA) module to identify auxiliary samples with high-uncertainty labels (\ie the slash star) and reduce their impact when aligning the representations at the semantic level.}
%   \label{fig:intro}
% \end{figure}

The above findings motivate us to look for new auxiliary data sources to facilitate medical image analysis. 
In the field of cervical dysplasia visual inspection, we observe that multiple image datasets exist (\eg NHS~\cite{Herrero2000PopulationbasedSO} and ALTS~\cite{Walker2003ART}), which are relevant but differ significantly in their collection environment. % and label quality. 
Intuitively, the knowledge learned from one dataset (\eg ALTS) will be helpful to improve the robustness of a model trained on another dataset (\eg NHS), which is, however, ignored by previous methods. %in this field ignore the relevant data sources and learn from a single target dataset only.
We also observe that a direct utilization of existing domain adaptation/generalization methods performs unsatisfactorily due to not only 
(1) \emph{domain shift} --- datasets are collected using different devices in different environments; 
but also (2) \emph{criterion mismatch} --- 
%the category boundaries of different datasets may not precisely match due to the subjective variance 
the standards for ground-truth annotation can be different due to the subjective variance
--- the diagnosis was made by a single medical staff (\eg nurse, doctor) purely based on visual inspection without confirming laboratory tests. 
% but also (2) \emph{label uncertainty}---the labels of some datasets may be less reliable together with subjective variance because the diagnosis was made by a single medical staff (\eg nurse, doctor) purely based on visual inspection without confirming laboratory tests. 

To tackle the above challenges, we present the first prototypical cross-domain knowledge transfer framework for cervical dysplasia visual inspection, which learns transferable information from an auxiliary dataset to improve the performance on the target dataset. 
As illustrated in Figure~\ref{fig:intro}, the framework has an edge in conducting simultaneous feature alignment under two distinct levels: domain level and class level. The \emph{Early Domain Alignment} (\emph{EDA}) module is presented to generate domain-aligned intermediate features, followed by the \emph{Prototypical Semantic Alignment} (\emph{PSA}) module producing semantically-consistent high-level representations across domains. 
Moreover, \emph{PSA} tackles the \emph{criterion mismatch} challenge by identifying and reducing the impact of the auxiliary samples with high-uncertainty labels. Specifically, \emph{PSA} first computes the class prototypes (\ie the feature centroid of each class) in the target domain as the reference to generate soft assignments for auxiliary samples. Next, it measures the cross-domain label consistency by comparing the soft assignments with the ground-truth labels of the auxiliary samples.
By thresholding the consistency score, we select reliable auxiliary samples and apply the supervised contrastive loss to pull together samples of the same class and push apart samples of different classes in the shared semantic space. Thereafter, semantically-consistent representations are learned across domains, which brings significant benefits for model optimization and knowledge transfer from the auxiliary to the target domain.
Here we summarize the key contributions of this paper as follows:

\begin{itemize}[leftmargin=*]
  \item To the best of our knowledge, we present the first cross-domain cervical dysplasia visual inspection method, which effectively transfers knowledge from the auxiliary to the target domain. %To reduce the domain gap, we align the features of domain-private encoders based on adversarial training before passing them to the high-level shared encoder.
  %We propose to align the intermediate features on \emph{domain level} with the existing modules and \emph{class level} with our \emph{PSA} module to facilitate model optimization.
  We propose to simultaneously align the intermediate features on both \emph{domain level} and \emph{class level} to learn transferable representations across domains.
%   \item We propose a novel prototype-based knowledge filtering method for inter-domain supervision. Images in the auxiliary dataset are either filtered out or supervised by cross-entropy or supervised contrastive loss, depending on their label transferability estimated based on the distance to the class prototypes in the target domain. 
  %\textcolor{blue}{ \item We propose a novel prototype-based semantic alignment method for class-level feature alignment. Criterion-similar images in the auxiliary dataset are selected for feature alignment using the supervised contrastive loss, depending on their label transferability estimated based on the distance to the class prototypes in the target domain.}
  \item We propose a novel prototype-based method to estimate the transferability of samples in the auxiliary domain. The impact of inconsistent labels can thus be reduced by weighting the auxiliary samples according to their transferability estimated based on the distance to the class prototypes of the target domain.
  \item We have performed extensive experiments on three benchmark cervical image datasets. The experimental results show that our proposed method outperforms the state-of-the-art cervical cancer visual inspection methods by a significant margin.
  \item We have presented additional experiments on the Visda-2017 dataset. Results demonstrate the effectiveness of our method in general image analysis in addition to the cervical domain.
\end{itemize}

\section{Related Works}

\textbf{Cervical Dysplasia Visual Inspection.}
A significant number of machine-learning-based methods for cervical dysplasia visual inspection have been proposed in recent years~\cite{Chang2005CombinedRA,DeSantis2007-SpectroscopicIA,MultimodalEC-Song2015,Xu2016MultimodalDL,ou2020semi}. 
%Apart from some studies that focused on the information integration from metadata~\cite{DeSantis2007-SpectroscopicIA,Xu2016MultimodalDL}, many methods are still using information purely from the image data.
%Different network architectures have been investigated to extract features from image data with the help of transfer learning~\cite{Vasudha2018CervixCC,chandran2021diagnosis,park2021comparison}. 
CYENet~\cite{chandran2021diagnosis} and ColpoNet~\cite{saini2020colponet} were network architectures tailored for cervical cancer detection with cross-norm operations. 
Zhang~\etal~\cite{zhanga2021spatial,chae2022attention} introduced a split-and-aggregation framework to process the high-resolution cervical images and provided classification results by summarizing patch features.
One alternative solution to leverage the high-resolution input is to train a cervix detector to generate the region of interest from the original image.
Faster-RCNN~\cite{Ren2015FasterRCNN} was adopted by Hu~\etal~\cite{Hu2019AnOS} as the detector, which was trained based on their self-annotated bounding box labels.
%Alyafeai~\etal~\cite{ALYAFEAI2020autopipeline} proposed a more general pipeline for detector training. Anyone with or without private cervix datasets can follow this pipeline for detector training using a public dataset.
Alyafeai~\etal~\cite{ALYAFEAI2020autopipeline} proposed a more general pipeline for detector training, following which cervical detectors can be trained using a public dataset.
%, which is also the pipeline we adopted in our experiments. 
Park~\etal~\cite{park2021comparison} applied multiple augmentation schemes to the cropped images, together with a ResNet-50 structure initialized with an ImageNet pre-trained model.
Some studies focused on the information integration from metadata such as Pap results~\cite{DeSantis2007-SpectroscopicIA} and HPV signal strength~\cite{Xu2016MultimodalDL}. However, only a small number of cervical images are associated with such metadata, which significantly limits the feasibility of such approaches.

\begin{figure*}[!t]
  \centering
  \includegraphics[width=.7\linewidth]{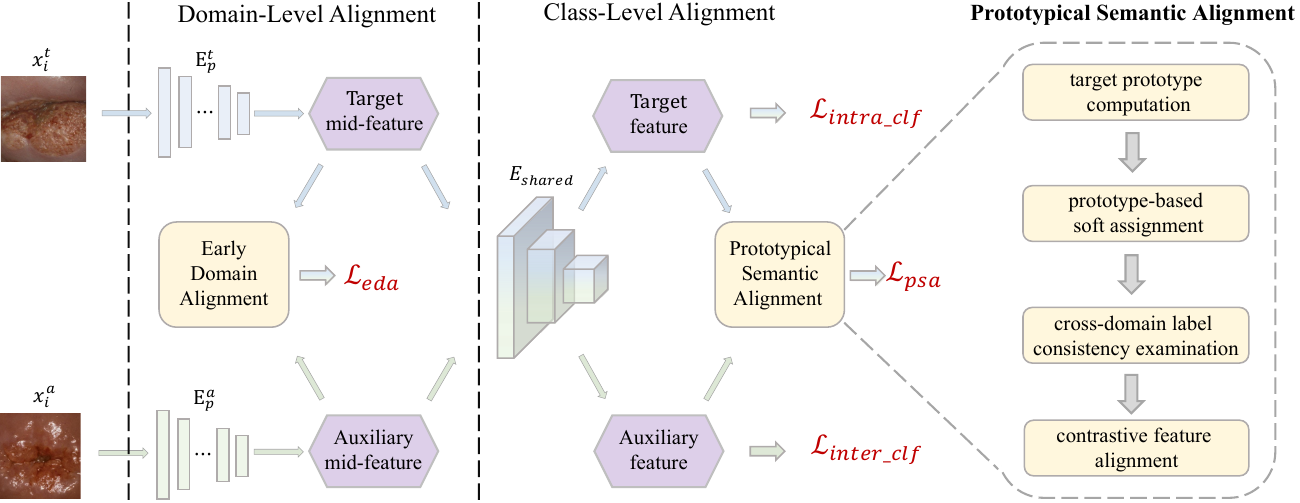}
  \caption{The overall architecture of our proposed Prototypical Cross-domain Knowledge Alignment and Transfer. $E_{p}$ and $E_{shared}$ denote the domain-private encoder and the shared encoder, respectively. }
  \label{fig:framework}
\end{figure*}

\textbf{Domain Adaptation. }
Domain adaptation (DA) focuses on transferring label information from the source domain to the target domain.
% One common idea is to perform cross-domain feature alignment.
% Previous methods~\cite{Pan2011DomainAV,Gong2013ConnectingTD,Long2015LearningTF,long2017deep} used feature-based methods to match the feature distribution across domains. %MK-MMD and JMMD was proposed by \cite{Long2015LearningTF} and \cite{long2017deep} as statistical criterion to measure the alignment progress.
Existing DA methods achieved it mainly based on 
statistical metrics~\cite{ ghifary2014domain,Long2015LearningTF,sun2016deep,long2017deep,zellinger2017central,peng2018synthetic}, 
semantic clustering~\cite{ zhang2015deep,saito2017asymmetric,motiian2017unified,yin2021enhanced,berthelot2021adamatch,Park2020JCL,harary2022unsupervised}, 
adversarial learning~\cite{ ganin2016domain,saito2018maximum,saito2019semi,CVPR2019MotionPrediction,chen2019progressive,jiang2020implicit,du2021cross}, 
or reconstruction~\cite{
liu2016coupled,Bousmalis2016DomainSN,CVPR2021DualConsecutiveNetwork,Cao2018DiDADS,yang2020label,peng2019domain}.
For example, CCSA~\cite{motiian2017unified} matched the cross-domain semantic space by aligning features based on their labels. DSN~\cite{Bousmalis2016DomainSN} proposed a disentanglement-based complex framework to separate style and content information. BrAD~\cite{harary2022unsupervised} designed an auxiliary bridge to narrow the gap between different domains. JCL~\cite{Park2020JCL} adopted a MoCo-like~\cite{He2020MoCo} structure to align unlabeled data and PAC~\cite{mishra2021surprisingly} introduced a pre-training stage for model training.
Since our goal differs from the DA task but shares similar properties, we select some of the existing works for comparison in our experiment. However, the asymmetrical designs of DA methods make them inappropriate to deploy in our setting, leading to worse performance compared to our framework.
% Other methods~\cite{Cao2018DiDADS,Yang2019UnsupervisedDA} adopted adversarial learning to achieve this goal, where the feature encoders try to fool the domain discriminators about the domain of input samples.

\textbf{Contrastive Learning.}
Contrastive learning was initially proposed to learn high-quality representations in a self-supervised manner where the positive pairs are constructed as the multiple augmentation views of the same sample~\cite{He2020MoCo,Chen2020MoCov2,Chen2020SimCLR,Chen2020SimCLRv2}. 
%As an important part of self-supervised learning, contrastive learning aims to make a model succeed in distinguishing (contrasting) between positive and negative pairs. 
%Various methods~\cite{He2020MoCo,Chen2020MoCov2,Chen2020SimCLR,Chen2020SimCLRv2} have been developed for unsupervised contrastive learning based on an instance discrimination task, where positive pairs are constructed as the multiple augmentation views of the same sample. 
%The key idea behind this scheme is narrowing the gap between positive pair representation, while pushing away the negative one.
For example, MoCo~\cite{He2020MoCo} proposed to use a momentum encoder and a large dictionary to improve model stability. %a novel framework that largely boosts the improvement of contrastive learning based on the momentum update and large dictionary.
SimCLR~\cite{Chen2020SimCLR} adopted a non-linear projection head to calculate the NT-Xent loss within the latent space.
Recently, contrastive learning has also been investigated under the supervised configuration, where positive pairs are defined as the same-class samples in a mini-batch~\cite{Khosla2020SupConLoss}. Multiple positive pairs are considered jointly in the calculation. 
%With the semantic context, we found that it can serve as an essential component for the cross-domain feature alignment.
In this paper, we further investigate supervised contrastive learning in a cross-domain setting for feature alignment.

\section{Problem Formulation}

The cervical dysplasia visual inspection is usually formulated as an image classification problem based on the CIN grades (CIN0 $\sim$ CIN4). Such an AI medical system can behave as a useful and efficient tool in alerting potential patients to take further medical examinations in real life, especially in low-resource regions where medical resources are deficient. 
%In the past decades, many deep learning based methods~\cite{Xu2017MultifeatureBB,Vasudha2018CervixCC,ALYAFEAI2020autopipeline,zhanga2021spatial,chandran2021diagnosis,park2021comparison} have been proposed for cervical dysplasia visual inspection. However, their model performance and robustness are generally limited by small-scale cervical datasets. 
However, the performance of existing deep learning models for cervical dysplasia visual inspection is generally limited by small-scale cervical datasets.
Moreover, the integration of multiple datasets will possibly lead to even worse performance if the aforementioned challenges of \emph{domain shift} and \emph{criterion mismatch} are not properly addressed.
Following this path, we focus on leveraging data from two different but relevant datasets (domains\footnote{We use these two terms interchangeably in this paper.}) to perform a more robust cervical dysplasia visual inspection. 
Given a target domain $X_t=\{x_1^t, x_2^t, \dots, x_{N^t}^t \}$ with labels $Y_t=\{y_1^t, y_2^t, \dots, y_{N^t}^t \}$ and an auxiliary domain $X_a=\{x_1^a, x_2^a, \dots, x_{N^a}^a \}$ with labels $Y_a=\{y_1^a, y_2^a, \dots, y_{N^a}^a \}$, our goal is to improve the performance of the model on the target domain $X_t$ with the facilitation of the auxiliary domain $X_a$. 

%Although domain adaptation methods~\cite{Bousmalis2016DomainSN,Park2020JCL} can be adopted to reduce \emph{domain shift} by extracting general features across domains, 
% our task is more challenging since the labels of our auxiliary domain can be less reliable than those of the target domain. 
% To jointly solve the \emph{domain shift} and \emph{label uncertainty} challenges, we propose a novel prototypical cross-domain knowledge transfer framework, which simultaneously aligns feature distributions at the domain level and improves feature discriminability at the class level.
Recall that the annotation quality of our auxiliary domain may not meet the standard of the target domain due to the \emph{criterion mismatch} challenge. The auxiliary labels $Y_a$ cannot be directly used for training.
We thus propose a novel prototypical cross-domain knowledge alignment and transfer framework. Without loss of generality, we sample $|S^t|$ and $|S^a|$ images from $X_t$ and $X_a$ in each iteration, where $S^t$ and $S^a$ represent the target domain mini-batch and the auxiliary domain mini-batch, respectively. Next, we introduce our proposed model architecture and optimization objective based on $S = S^t \cup S^a$ in each iteration.

%\textcolor{blue}{Let $S^t$ denotes the target mini-batch and $x^t \in S^t$ is the target images. Then for each iteration, we sample $|S^t|$ images from the target and auxiliary domain, respectively.}
%which aligns the feature distributions at both domain level and class level. By optimizing the shared semantic space, the alignment can provide a more comprehensive horizon during model training.

\section{Prototypical Cross-domain Knowledge Alignment and Transfer}
The architecture overview of our proposed prototypical cross-domain knowledge alignment and transfer framework is illustrated in Figure~\ref{fig:framework}.
As aforementioned, our framework consists of an Early Domain Alignment (\emph{EDA}) module for domain-level feature alignment and a Prototypical Semantic Alignment (\emph{PSA}) module for class-level feature alignment. The \emph{PSA} module further estimates the transferability of the auxiliary samples to reduce the impact of label inconsistency. By jointly optimizing the feature alignment and the classification objectives, cross-domain transferable knowledge can be effectively learned and transferred to the target domain.

\subsection{Early Domain Alignment}
Intuitively, a shared encoder is preferred for performance improvement if we try to introduce auxiliary data for training. 
However, different domains are generally occupied with differences in local feature distributions. Therefore, we adopt a Y-shape domain-adapted architecture as illustrated in Figure~\ref{fig:framework} to deal with the \emph{domain shift}. It consists of two domain-private encoders ($E_p^t$ and $E_p^a$) for local information extraction, a shared encoder for high-level semantic deduction ($E_{shared}$). 
On top of that, an early domain alignment module is introduced to reduce the gap between the intermediate representations extracted by the two domain-private encoders. 

To obtain domain-invariant features, we have investigated two major approaches to narrow the gap between domains:

\textbf{Adversarial-based.}
Following \cite{ganin2016domain}, the adversarial-based alignment is achieved by minimizing the domain classification loss for the domain classifier $g$, while maximizing this loss for the encoders, with the help of a gradient reversal layer. We formulate the adversarial-based objective for early domain alignment as
\begin{equation}
\small
    \mathop{\max}_{E_p}\mathop{\min}_{g} \mathcal{L}_{eda} = - \frac{1}{|S|} \sum_{i=1}^{|S|} (y^d_i\log(\hat{y}^d_i) + (1 - y^d_i)\log(1 - \hat{y}^d_i)),
    \label{eqn:eda_adv}
\end{equation}
where $\hat{y}^d_i$ is the output of the domain classifier, and $y^d_i$ is the domain index of the input image (\ie 0 for target and 1 for auxiliary). 

\textbf{Divergence-based.}
An alternative for domain-level alignment is divergence-based approaches, the goal of which is to minimize the distance between representations of samples from different domains. Here we investigate a widely used distance metric termed MK-MMD~\cite{Long2015LearningTF}.
%\begin{equation}
%\small
%    MKMMD(x^t,x^a) = ||\mathbf{E}^t [ \phi(x^t)] - \mathbf{E}^a [\phi(x^a)] ||^2_{\mathcal{H}_k}
%    \label{eqn:mkmmd}
%\end{equation}
Thereby, the divergence-based objective is to minimize the $MKMMD$ distance between the intermediate features of $x^t$ and $x^a$ to narrow the gap across domains:
\begin{equation}
\small
    \mathop{\min}_{E_p} \mathcal{L}_{eda} = MKMMD(GAP(E_p^t(x_i^t)),GAP(E_p^a(x_i^a))),
    \label{eqn:eda_diver}
\end{equation}
where $GAP$ is the global average pooling that maps the output of the private encoders $E^t_p$ and $E^a_p$ into a vector.

\begin{figure}[!t]
  \centering
  \includegraphics[width=.8\linewidth]{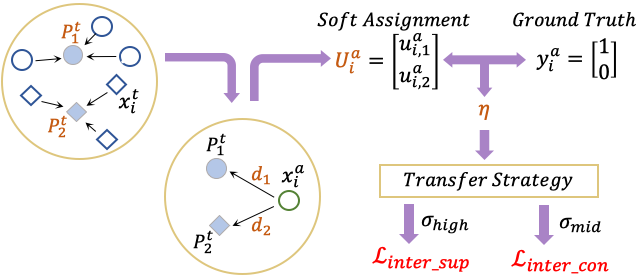}
  \caption{The pipeline of proposed PSA module.}
  \label{fig:filter}
\end{figure}

\subsection{Prototypical Semantic Alignment}
% Here we introduce how to learn and transfer useful information from an auxiliary domain. Similarly, the high-level representations of auxiliary images extracted by the shared encoder $E_{shared}$ are fed to a shared dense layer followed by a private \emph{PDKF} module to perform inter-domain supervision.
% Recall that in our auxiliary dataset, labels are generally provided by a single medical staff from a local clinic. The large variance of technical skills and subjective deviations lead to the \emph{label uncertainty} challenge. 
% To solve this issue, we present a \emph{Prototypical Domain Knowledge Filtering} (\emph{PDKF}) module to perform effective knowledge transfer from the auxiliary domain. As shown in Figure~\ref{fig:filter}, it consists of target prototype computation, prototype-based soft assignment, cross-domain label consistency examination, and hierarchical knowledge transfer strategy.
%Here we introduce how to align class-level features between different domains. Similarly, the high-level representations of auxiliary images extracted by the shared encoder $E_{shared}$ are fed to a shared dense layer followed by a private \emph{PDKA} module to perform class-level feature alignment.
Recall that in our auxiliary dataset, labels are generally provided by a single medical staff from a local clinic. The large variance of technical skills and subjective deviations lead to the \emph{criterion mismatch} challenge. 
To reduce the impact of such label inconsistency, we present a \emph{Prototypical Semantic Alignment} (\emph{PSA}) module to align the semantics of the high-level feature representations (\ie output of $E_{shared}$) between the target and auxiliary domains. As shown in Figure~\ref{fig:filter}, it consists of target prototype computation, prototype-based soft assignment, cross-domain label consistency examination, and contrastive feature alignment.

\textbf{Target Prototype Computation.}
We propose to compute the per-class prototypes in the target domain, and use them as references to deduce reliable and transferable information from the auxiliary domain. A prototype is defined as the center of a semantic cluster, consisting of features with the same semantic label. 
Compared with instance-to-instance matching~\cite{motiian2017unified,kim2020cross}, where matching is performed between cross-domain instance pairs, instance-to-prototype matching is more robust to abnormal instances, thus providing a better foundation for the following steps.

Specifically, we append a classification head that consists of two fully-connected layers (\ie $FC_1$ and $FC_2$) on top of the shared encoder,
and compute the prototype of each target class $k$ after every epoch as the average of all the features from this class by
%More specifically, the prototype of each target class $k$ is computed after every epoch as the average of all the features from this class by
\begin{equation}
    \small
    P^t_{k} = \frac{\sum_{i=1}^{N^t} f^t(x^t_i)\mathds{1}_{(y^t_i==k)}}{\sum_{i=1}^{N^t} \mathds{1}_{(y^t_i==k)}}, 
\end{equation}
where $f^t(x^t)=FC_1(E_{shared}(E_p^t(x^t)))$ maps target image $x^t$ to the feature before the last classification layer. $P^t_{k}\in\mathbb{R}^{256} $ represents the prototype of class $k$, and $N^t=|X_t|$ is the number of target training samples. 
The prototypes are denoted as $P^t=[P^t_1, P^t_2, \ldots, P^t_K]$, where $K$ is the total number of classes.

%\textcolor{red}{what is the relation between f(x), projection head, and classification head? sorry I don't recall this implementation detail...}
%\textcolor{blue}{(projection head and classification head are built on top of the convolutional encoder. The Linear here is the 1st FC layer of the classification head.)}
\textbf{Prototype-based Soft Assignment.}
The target prototypes $P^t$ are next utilized to calculate a distance-based soft assignment for each auxiliary sample as follows
\begin{equation}
    \small
    u^a_{i,k} = \frac{exp(-||f^a(x^a_i)-P_k^t||_2)}{\sum_{c=1}^{K} exp(-||f^a(x^a_i)-P_c^t||_2)},
\end{equation}
where $U^a_{i}=[u^a_{i,1},u^a_{i,2},\ldots,u^a_{i,K}]\in\mathbb{R}^{K}$ is a K-dimensional vector representing the probability of $x^a_i$ belonging to target class $k$. $f^a(x^a)=FC_1(E_{shared}(E_p^a(x^a)))$ maps auxiliary image $x^a$ to the shared feature space, $||f^a(x^a_i)-P_k^t||_2$ computes the $l_2$ distance between $f^a(x^a_i)$ and $P_k^t$, and softmax is applied to normalize $U^a_{i}$ by ensuring $\sum_k u^a_{i,k}=1$. Thereby, an auxiliary sample $x^a_i$ will be assigned with a large $u^a_{i,k}$ if it is close to the prototype of class $k$ in the shared feature space.

\textbf{Cross-domain Label Consistency Examination.}
For an auxiliary sample $x^a_i$ with label $y^a_i=k$, it should be close to $P^t_k$ in the feature space if it is well aligned with the target domain. Based on this observation, we propose to compute a cross-domain label consistency score for each auxiliary sample to measure the reliability of the knowledge learned from it.
Given the calculated soft assignment probabilities $U^a_{i}$ and the original auxiliary ground truth $y^a_i$, the cross-domain label consistency is formally defined as
\begin{equation}
    \small
    \eta_{i} = [U^a_{i}]^\top \cdot y^a_i,
\label{eq:eta}
\end{equation}
where $\eta_{i}\in[0,1]$ is the consistency score, and operator $\cdot$ represents the dot product of two vectors. The closer $\eta_{i}$ is to $1$, the higher the confidence that this auxiliary sample is within the same decision boundary of the target domain.
\begin{table*}[t!]
%\small
\centering
% \fontsize{9}{11}\selectfont
\caption{Performance comparison between our method and state-of-the-art methods on the NHS dataset.}

%\resizebox{\textwidth}{!}{
\begin{tabular}{c|c|c||ccc|c}
\Xhline{1.3pt}
Methods           & Aux data & Accuracy (\%) & Precision (\%) & Recall (\%) & F1 Score (\%) & ROC-AUC        \\
\hline\hline
% Vasudha’18~\cite{Vasudha2018CervixCC}        &   \xmark    & 72.35$\pm$0.33   & 64.74$\pm$4.00    & 75.79$\pm$14.65 & 69.04$\pm$4.64   & 0.7716$\pm$0.0119  \\
% Alyafeai'20-2conv~\cite{ALYAFEAI2020autopipeline} &    \xmark      & 72.91$\pm$2.15   & 64.02$\pm$1.20    & 78.99$\pm$6.76 & 70.66$\pm$3.49   & 0.7813$\pm$0.0054 \\
% CYENet~\cite{chandran2021diagnosis}            &    \xmark      & 76.51$\pm$1.74   & 66.94$\pm$1.99    & 85.84$\pm$3.45 & 75.19$\pm$1.86  & 0.7773$\pm$0.0200  \\
2conv-Alyafeai~\etal~\cite{ALYAFEAI2020autopipeline} &    \xmark      & 71.02$\pm$2.15   & 64.10$\pm$1.20    & 68.49$\pm$6.76 & 66.22$\pm$3.49   & 0.7327$\pm$0.0054 \\
3conv-Alyafeai~\etal~\cite{ALYAFEAI2020autopipeline} &   \xmark      & 72.16$\pm$1.50   & 62.24$\pm$2.30    & 83.56$\pm$2.37 & 71.34$\pm$1.16   & 0.7442$\pm$0.0143  \\
CYENet~\cite{chandran2021diagnosis}            &    \xmark      & 75.57$\pm$1.74   & 72.06$\pm$1.99    & 67.12$\pm$3.45 & 69.50$\pm$1.86  & 0.7961$\pm$0.0200  \\
Vasudha~\etal~\cite{Vasudha2018CervixCC}        &   \xmark    & 77.27$\pm$0.33   & 67.74$\pm$4.00    & 81.30$\pm$9.65 & 75.90$\pm$4.64   & 0.7885$\pm$0.0119  \\
% Alyafeai'20-3conv~\cite{ALYAFEAI2020autopipeline} &   \xmark      & 76.70$\pm$1.50   & 67.46$\pm$2.30    & 84.93$\pm$2.37 & 75.15$\pm$1.16   & 0.7911$\pm$0.0143  \\
Zhang~\etal~\cite{zhanga2021spatial}          &    \xmark     & 81.82$\pm$1.14   & 73.38$\pm$1.19    & 88.12$\pm$1.58 & 80.07$\pm$1.26   & 0.8351$\pm$0.0071  \\
% Park'21~\cite{park2021comparison}           &    \xmark      & 82.84$\pm$1.29   & 79.48$\pm$4.10    & 79.72$\pm$5.92 & 79.36$\pm$1.52   & 0.8275$\pm$0.0153  \\
%ResNet-50~\cite{He2016resnet}           &    \xmark      & 82.84$\pm$1.29   & 79.48$\pm$4.10    & 79.72$\pm$5.92 & 79.36$\pm$1.52   & 0.8275$\pm$0.0153  \\
\hline
% JCL 0.8125(15), Prec: 0.7778, Recall: 0.7671, F1: 0.7724 AUC: 0.8465
% CCSA 0.8182 Precision 0.7356 Recall 0.8767 F1 Score 0.7999 Auc 0.8564
% DSN 0.8239 Precision 0.7917 Recall 0.7808 F1 Score 0.7862 AUC 0.8577 

% DSN~\cite{Bousmalis2016DomainSN}               &  \cmark & 81.62$\pm$1.43   & 77.28$\pm$4.32    & 79.45$\pm$3.62 & 78.21$\pm$0.60   & 0.8499$\pm$0.0249  \\
% DSN+aux label~\cite{Bousmalis2016DomainSN}     &  \cmark & 82.19$\pm$1.18   & 77.42$\pm$3.87    & 81.27$\pm$8.26 & 79.01$\pm$2.39   & 0.8547$\pm$0.0204  \\
PAC~\cite{mishra2021surprisingly} &  \cmark  & 79.55$\pm$1.01 & 71.33$\pm$0.92 & 84.96$\pm$3.23 & 77.51$\pm$1.16 &  0.8411$\pm$0.0179   \\
BrAD~\cite{harary2022unsupervised}              &  \cmark & 81.25$\pm$1.20  & 76.32$\pm$5.91    & 79.45$\pm$7.75 & 77.85$\pm$0.29   & 0.8256$\pm$0.0031  \\
JCL~\cite{Park2020JCL}               &  \cmark & 81.39$\pm$0.98   & 75.64$\pm$2.58    & 80.82$\pm$2.74 & 78.14$\pm$0.52   & 0.8493$\pm$0.0124  \\
CCSA~\cite{motiian2017unified}              &  \cmark & 81.65$\pm$1.14   & 73.56$\pm$1.36    & 87.67$\pm$1.58 & 79.99$\pm$1.33   & 0.8564$\pm$0.0052  \\
% PAC~\cite{mishra2021surprisingly} &  \cmark  & 82.21$\pm$1.01 & 76.92$\pm$0.92 & 82.21$\pm$3.23 & 79.51$\pm$1.16 &  0.8598$\pm$0.0179   \\
DSN~\cite{Bousmalis2016DomainSN}               &  \cmark & 82.38$\pm$1.43   & 79.17$\pm$4.32    & 78.08$\pm$3.62 & 78.62$\pm$0.60   & 0.8577$\pm$0.0249  \\
% JCL~\cite{Park2020JCL}               &  \cmark & 82.95$\pm$0.98   & 75.40$\pm$2.58    & 87.67$\pm$2.74 & 81.01$\pm$0.52   & 0.8535$\pm$0.0124  \\
% BrAD~\cite{harary2022unsupervised}              &  \cmark & 83.24$\pm$1.20  & 77.51$\pm$5.91    & 84.93$\pm$7.75 & 80.77$\pm$0.29   & 0.8547$\pm$0.0031  \\
% JCL+aux label~\cite{Park2020JCL}      &  \cmark & 83.52$\pm$0.98   & 77.95$\pm$4.87    & 84.93$\pm$7.63 & 80.98$\pm$1.50   & 0.8657$\pm$0.0145  \\
% CCSA~\cite{motiian2017unified}              &  \cmark & 83.52$\pm$1.14   & 78.45$\pm$1.36    & 83.10$\pm$1.58 & 80.70$\pm$1.33   & 0.8714$\pm$0.0052  \\
\hline
\textbf{Ours-mkmmd}  &  \cmark & 85.80$\pm$0.57   & 79.27$\pm$0.87    & 84.04$\pm$3.72 & 83.87$\pm$1.18   & \textbf{0.8832$\pm$0.0093} \\
\textbf{Ours-adv} &  \cmark & \textbf{86.55$\pm$0.88}   & \textbf{80.42$\pm$2.09}    & \textbf{89.49$\pm$2.09} & \textbf{84.67$\pm$0.78}   & 0.8822$\pm$0.0111 \\
\Xhline{1.3pt}
\end{tabular}
%}
\label{table:sota_nhs}
\end{table*}

\textbf{Contrastive Feature Alignment.}
%Based on a predefined confidence threshold $\sigma_{align}$, we select reliable auxiliary samples for cross-domain semantic alignment according to the consistency score $\eta_{i}$ calculated using Eqn.~\ref{eq:eta}.
%More specifically, samples with consistency scores $\eta_{i}\ge\sigma_{align}$ stay within the tolerable range of criterion dissimilarity and are thus utilized for the shared semantic space optimization. 
Given the consistency score $\eta_{i}$ calculated using Eqn.~\ref{eq:eta} and a predefined threshold $\sigma_{align}$, we filter auxiliary samples by only keeping those with consistency scores $\eta_{i}\ge\sigma_{align}$ to align the high-level features. Following SimCLR~\cite{Chen2020SimCLR}, we apply a projection head on top of the shared encoder and perform supervised contrastive learning in the projection space~\cite{Khosla2020SupConLoss}. By pulling together samples of the same class and pushing apart samples of different classes in the projection space, it introduces consistent performance gain for classification models.
In our implementation, positive pairs $(x_i,x_j)$ are defined as images that belong to the same semantic class (\ie $y_i=y_j$), while negative pairs are images that belong to different semantic classes (\ie $y_i \neq y_j$). 
Both cross-domain (\ie $(x^t_i,x^t_j)$ or $(x^a_i,x^a_j)$) and intra-domain $(x^t_i,x^a_j)$ pairs are jointly considered to compute the supervised contrastive loss $\mathcal{L}_{sc}$. For simplicity, we omit the superscript in the formulation of $\mathcal{L}_{sc}$, which is given as
\begin{equation}
\small
    \mathcal{L}_{sc}(x_i) = - \log  \frac{1}{|Q^+(i)|} \sum_{x_p} \frac{exp(sim(\mathbf{z}_i, \mathbf{z}_p))}{\sum\limits_{x_q} exp(sim(\mathbf{z}_i , \mathbf{z}_q)) },  
\label{supcon_loss}
\end{equation}
where $\mathbf{z}_i$ is the output of the projection head corresponding to $x_i$, and $sim()$ computes the cosine similarity.
$x_q \in Q(i)\equiv\{I-x_i\}$ where $I$ is the union of the target batch and the filtered auxiliary batch based on $\sigma_{align}$.
%$B$ and $\sum_{i=1}^{B} \mathbbm{1}_{(\eta_{i}\ge\sigma_{align})}$ samples from the target and auxiliary domains, respectively. 
$x_p \in Q^+(i) \equiv \{x_q: y_q = y_i\}$ contains all the positive pairs $(x_i, x_p)$ for $x_i$ in $Q(i)$. 
%Following SimCLR~\cite{Chen2020SimCLR}, a projection head with output $\mathbf{z}_i$ is applied on top of the shared encoder to perform contrastive learning. 
The objective is to maximize the cosine similarity between positive pairs while minimizing it between negative pairs. Subsequently, the prototypical semantic alignment loss is computed as the average of the supervised contrastive loss over all valid training samples:

\begin{equation}
\small
    % \mathcal{L}_{psa} = \frac{\sum_{i=1}^{B} \mathcal{L}_{sc}(x^t_i) + \sum_{i=1}^{B} \mathbbm{1}_{(\eta_{i}\ge\sigma_{align})} \mathcal{L}_{sc}(x^a_i)}
    % {B+\sum_{i=1}^{B} \mathbbm{1}_{(\eta_{i}\ge\sigma_{align})}},
    \mathcal{L}_{psa} = \frac{\sum_{i=1}^{|S^t|} \mathcal{L}_{sc}(x^t_i) + \sum_{i=1}^{|S^a|} \mathds{1}_{(\eta_{i}\ge\sigma_{align})} \mathcal{L}_{sc}(x^a_i)}
    {|S^t|+\sum_{i=1}^{|S^a|} \mathds{1}_{(\eta_{i}\ge\sigma_{align})}}.
\label{psa_loss}
\end{equation}

Our proposed prototypical semantic alignment loss strengthens the matching of cross-domain samples. Considering that the visual appearances of cervical images are highly similar across cases, it can also assist in quick concentration on the most important information for classification.

\subsection{Cross-Domain Knowledge Transfer}
%\textcolor{red}{Eqn 1 and 5, we compute average loss, here in Eqn 6, we compute sum? It should also be divided by $|B|$ right? Also in Eqn 1, should it be divided by 2$|B|$?}
%\textcolor{blue}{(Yes, but 2B in previous sections might make people confused. And it's also correct to compute loss for each domain separately. No need to integrate into the same equation. So it's like using the Eqn 1 twice.)}
The aforementioned feature alignments assist our model to generate domain-invariant features with a global horizon. 
%Based on these features, we apply a classification head on top of the shared encoder for classification.
%After offering global model horizon via feature alignment, 
Next, we perform cross-domain knowledge transfer based on supervised classification.
We adopt a threshold $\sigma_{clf}$ for cross-domain knowledge transfer with respect to the consistency score computed in Eqn.~\ref{eq:eta}. 
Let $x^t \in S^t$ and $x^a \in \tilde{S}^a=\{x^a_i|\eta_{i}\ge\sigma_{clf}\}$ denote the training samples in the target mini-batch and in the filtered auxiliary mini-batch based on $\sigma_{clf}$, respectively, then the cross-entropy for classification is computed as
% We conduct the intra-domain supervision based on target images and labels via the cross-entropy loss as
\begin{equation}
    \small
    % \mathcal{L}_{intra\_sup} = - \frac{1}{B} \sum_{i=1}^{B} y^t_{i}\log(\hat{y}^t_{i}),
    \begin{split}
    \mathcal{L}_{clf} &= \mathcal{L}_{intra\_clf} + \gamma \mathcal{L}_{inter\_clf} \\
    &= - \frac{1}{|S^t|} \sum_{x^t_i\in S^t} y^t_i  \log(\hat{y}^t_i) - \frac{\gamma}{|\tilde{S}^a|} \sum_{x^a_i\in \tilde{S}^a} \eta_i y^a_i  \log(\hat{y}^a_i),
    % &= - \frac{1}{B} \sum_{}^{} y^t_i  \log(\hat{y}^t_i) - \frac{\gamma}{\mathbbm{1}_{(\eta_{i}\ge\sigma_{clf})}} \sum_{}^{} \mathbbm{1}_{(\eta_{i}\ge\sigma_{clf})} \eta_i y^a_i  \log(\hat{y}^a_i),
    % &= \sum_{x_i\in\{S^t \cup S^a\}}^{} - \eta_i \gamma_i y_i  \log(\hat{y}_i),
    \end{split}
    \label{ce_loss_tar}
\end{equation}
where $\hat{y}_i$ and $y_i$ denote the prediction and the ground-truth label of image $x_i$, respectively. We further weight the auxiliary samples by $\gamma \cdot \eta_i$ to enforce stronger supervision from samples with a larger cross-domain consistency. %$\eta^t_i=\gamma^t_i=1$ for all target samples and $\gamma^s_i=10$ for auxiliary samples.
It serves as the mainstay of our framework, pushing it to constantly focus on extracting informative features for classification.

\subsection{Overall Objectives}
We optimize our model by jointly considering the classification loss, $\mathcal{L}_{clf}$, and the feature alignment losses, $\mathcal{L}_{eda}$ and $\mathcal{L}_{psa}$.
The overall loss function of our proposed prototypical cross-domain knowledge transfer framework is formulated as
\begin{equation}
\small
    % \mathcal{L} = \mathcal{L}_{intra\_sup} + \alpha \mathcal{L}_{ada} + \beta \mathcal{L}_{inter\_sup}  + \gamma \mathcal{L}_{inter\_con},
    \mathcal{L} = \mathcal{L}_{clf} + \alpha \mathcal{L}_{eda} + \beta \mathcal{L}_{psa},
    \label{loss_all}
\end{equation}
where $\alpha, \beta$ are coefficients controlling the balance between the classification loss and the feature alignment loss functions.

\section{Experiments}

\subsection{Dataset}

Totally 17,002 cervical images are used in our experiments, which were collected from three separate medical studies: Natural History Study of HPV and Cervical Neoplasia (NHS)~\cite{Herrero2000PopulationbasedSO}, ASCUS-LSIL Triage Study (ALTS)~\cite{Walker2003ART}, and Biopsy Study (Biopsy)~\cite{biopsy}.
We filter the records that are labeled with ground-truth CIN grades (CIN 0,1,2,3,4) within 1 year of the screening date and formulate it as a binary classification problem to detect abnormal cases, following previous work~\cite{zhanga2021spatial}.
The accessibility of these datasets is based on request and constrained agreement.
When compared to the state-of-the-art, we performed two sets of experiments by utilizing NHS and Biopsy as the target dataset, respectively, while the NHS dataset is utilized in ablation studies.
Please refer to the supplementary material for a detailed description of these datasets.

\begin{table*}[t!]
\centering
% \fontsize{9}{11}\selectfont
\caption{Performance comparison between our method and domain adaptation methods on the Biopsy dataset.}

\begin{tabular}{c|c|c||ccc|c}
\Xhline{1.3pt}
Methods       & Aux data                   & Accuracy (\%) & Precision (\%) & Recall (\%)  & F1 Score (\%)                   & ROC-AUC       \\
\hline\hline
% Park'21~\cite{park2021comparison}           &    \xmark      & 62.30$\pm$0.00   & 64.61$\pm$0.86    & 51.67$\pm$2.35 & 57.39$\pm$1.12   & 0.6038$\pm$0.0069  \\
%ResNet-50~\cite{He2016resnet}           &          & 62.30$\pm$0.21   & 64.61$\pm$0.86    & 51.67$\pm$2.35 & 57.39$\pm$1.12   & 0.6038$\pm$0.0069  \\
% cyenet &0.5574 Precision 0.5455 Recall 0.6000 F1 Score 0.5714 Auc 0.5280    \\
% auto-2conv &0.5574 Precision 0.5652 Recall 0.4333 F1 Score 0.4905 Auc 0.5505  \\
% auto-3conv &0.6066 Precision 0.5937 Recall 0.6333 F1 Score 0.6129 Auc 0.5677 \\
% lenet &0.6066 Precision 0.6875 Recall 0.3667 F1 Score 0.4782 Auc 0.6301 \\
% \hline
CCSA~\cite{motiian2017unified}          &  \cmark & 56.56$\pm$1.16   & 66.27$\pm$7.30    & 26.67$\pm$14.14 & 36.42$\pm$13.30                     & 0.5398$\pm$0.0471 \\
PAC~\cite{mishra2021surprisingly} &  \cmark  & 58.78$\pm$1.92 & 62.59$\pm$2.72 & 36.40$\pm$7.25 & 46.02$\pm$4.28 &  0.5511$\pm$0.0225   \\
BrAD~\cite{harary2022unsupervised}              &  \cmark & 59.02$\pm$0.06   & 57.99$\pm$1.83    & 62.22$\pm$11.70 & 59.54$\pm$4.66   & 0.5566$\pm$0.0022  \\ 
JCL~\cite{Park2020JCL}            &   \cmark & 59.84$\pm$1.08   & 61.93$\pm$1.73    & 48.33$\pm$11.79  & 53.79$\pm$6.82                     & 0.5801$\pm$0.0115 \\
% JCL+aux label~\cite{Park2020JCL} &   \cmark & 61.48$\pm$1.41   & 65.02$\pm$5.28    & 50.00$\pm$18.85 & 55.05$\pm$10.23                     & 0.6188$\pm$0.0373 \\
DSN~\cite{Bousmalis2016DomainSN}            &   \cmark & 62.29$\pm$1.64   & 68.03$\pm$9.20   & 47.77$\pm$9.62 & 55.11$\pm$4.45                    & 0.6208$\pm$0.0078 \\
% DSN+aux label~\cite{Bousmalis2016DomainSN}       &   \cmark & 62.84$\pm$2.50   & 63.66$\pm$5.28    & 58.88$\pm$5.09  & 60.89$\pm$0.47                     & 0.6050$\pm$0.0295 \\
\hline
\textbf{Ours-mkmmd}          &   \cmark & 63.93$\pm$0.92   & \textbf{72.22$\pm$3.29}    & 43.33$\pm$6.42 & 54.17$\pm$1.77 & \textbf{0.6269$\pm$0.0153} \\
\textbf{Ours-adv}          &   \cmark & \textbf{64.75$\pm$1.16}   & 63.48$\pm$0.22    & \textbf{66.66$\pm$4.72}  & \textbf{65.00$\pm$2.36} & 0.6253$\pm$0.0023 \\
\Xhline{1.3pt}
\end{tabular}
\label{table:sota_biopsy}
\end{table*}

% In the evaluation, we performed two sets of experiments by utilizing NHS and Biopsy as the target dataset, respectively. NHS is a widely adopted benchmark dataset in previous studies~\cite{Hu2019AnOS,zhanga2021spatial}, which is annotated by two experts in the National Testing Laboratory. Biopsy forms a more challenging small-scale dataset, which consists of 393 images only.
% ALTS is utilized as the auxiliary dataset in both cases. This dataset is annotated by different pathologists across clinics with varying professional levels, resulting in less reliable labels with subjective variance. %compared to the target datasets.

%In our experiments, we regard NHS as the target domain and ALTS as the auxiliary domain since NHS is a widely adopted benchmark dataset with smaller data scale in previous studies~\cite{Hu2019AnOS,zhanga2021spatial}, which is annotated by two experts in the National Testing Laboratory. The ALTS dataset, on the other hand, is annotated by different pathologists across clinics with varying professional levels, resulting in less reliable labels with subjective variance compared to NHS. Biopsy is a small-scale dataset being regarded as another target domain in a separate experiment for further validation.

\subsection{Implementation Details}

The original resolution of cervical images is generally 2,400$\times$1,600.
Following previous work~\cite{ALYAFEAI2020autopipeline}, we adopt a cropping scheme to select the region of interest as a preprocessing step. We adopt the ResNet-50~\cite{He2016resnet} as our backbone and initialize it with the ImageNet self-supervised model Dino~\cite{Caron2021EmergingPI}. %We append a projection head and a classification head on top of the shared encoder (\ie the last stage of ResNet-50) to perform contrastive learning and classification, respectively. Both the projection head and the classification head are implemented as two fully-connected layers with ReLU activation.
% We train our model using the Adam optimizer with weight decay set to $10^{-3}$.
% We adopt a mini-batch size of $|S^t|=|S^a|=128$ and an initial learning rate of $10^{-4}$. As the auxiliary domain can be much larger than the target domain, we find it to be beneficial by sampling balancedly from the two domains with a ratio of $1:1$. 
For training stability, we first train our model without the prototypical semantic alignment loss for 5 epochs as a warm-up, then continue training by empirically setting the balancing coefficients $\alpha, \beta, \gamma$ in the objective function to $0.1, 0.01, 0.1$, respectively. 
%The two-level thresholds were empirically set to $\sigma_{clf}=0.9$ and $\sigma_{align}=0.4$.
%For training stability, before applying the prototypical semantic alignment, we perform $5$ epochs as the warm-up stage with supervision based on image labels in both the target and auxiliary datasets.
We conduct an ablation study to evaluate the impact of the thresholds $\sigma_{align}$ and $\sigma_{clf}$ for cross-domain knowledge transfer, based on which we set $\sigma_{align}=0.4$ and $\sigma_{clf}=0.9$ in the rest of the experiments.
More details of implementation can be found in the supplementary material.

\subsection{Comparison to the State-of-the-Art}
We compare our proposed framework with nine state-of-the-art methods for both cervical dysplasia visual inspection and domain adaptation. 
Five commonly used classification measurements including top-1 accuracy, precision, recall, F1-score, and area under the ROC curve (ROC-AUC) are adopted as evaluation metrics. 
For a fair comparison, we train each model three times to reduce randomness, and report the average results together with the standard deviation of the three independent runs in Tables~\ref{table:sota_nhs} and~\ref{table:sota_biopsy}.

Table~\ref{table:sota_nhs} illustrates the performance comparison on the NHS dataset. Compared to previous cervical dysplasia visual inspection methods, our proposed framework with either divergence or adversarial alignment surpasses them by an overall large margin. Among the two candidates, the adversarial one performs better, which outperforms the second-best solution~\cite{zhanga2021spatial} by an average improvement of 4.73\% in top-1 accuracy, 7.04\% in precision, 1.37\% in recall, 4.60\% in F1 score, and 0.047 in ROC-AUC. 
%3.71\% in top-1 accuracy, 0.94\% in precision, 9.77\% in recall, 5.31\% in F1 score, and 0.055 in ROC-AUC. 
%Recall is considered to be a more important metric than precision as ideally all abnormal cases should be retrieved for further examination. 
A larger gap can be observed in various metrics against the rest of the methods~\cite{Vasudha2018CervixCC,ALYAFEAI2020autopipeline,chandran2021diagnosis}, where an improvement of more than 10\% in top-1 accuracy, 8\% in precision, 6\% in recall and 0.1 in ROC-AUC are generally obtained. 
The experimental results verify our motivation of looking for new auxiliary data sources for medical image analysis. Knowledge can be learned and transferred between medical images that were collected under different trials effectively, whereas the key challenges caused by domain shift and criterion mismatch across medical trials have to be properly solved.

Next, we compare our proposed framework to domain adaptation methods DSN~\cite{Bousmalis2016DomainSN}, JCL~\cite{Park2020JCL}, CCSA~\cite{motiian2017unified}, PAC~\cite{mishra2021surprisingly}, BrAD~\cite{harary2022unsupervised}, where auxiliary data are also utilized for training. 
% DSN and JCL were originally developed for unsupervised domain adaptation, we thus derived two additional variants from them to make use of the auxiliary labels by adding a cross-entropy loss in the objective function. 
% As can be seen, although the variants (\ie DSN+aux label and JCL+aux label) outperform their counterparts DSN and JCL, their improvements over the existing cervical dysplasia visual inspection methods are quite limited, especially in terms of top-1 accuracy and F1 score. 
% CCSA and BrAD outperform DSN and JCL on the NHS dataset. However, their performance degrade significantly on the Biopsy dataset (see Table~\ref{table:sota_biopsy}).
Originally designed for unsupervised domain adaptation (UDA) or domain generalization, these methods can be applied to our setting by considering our target domain as the target domain in UDA and our auxiliary domain as the source domain in UDA. This approach allows us to utilize both the target and auxiliary labels by adding a target cross-entropy loss on top of their corresponding objective function.
As can be seen, despite the fact that DSN performs the best out of all the existing methods on both datasets, its improvements over cervical dysplasia visual inspection methods are somewhat limited, particularly in terms of top-1 accuracy, recall, and F1 score.
We can also see that CCSA outperforms BrAD, PAC, and JCL on the NHS dataset. However, its performance degrades significantly on the Biopsy dataset (refer to Table~\ref{table:sota_biopsy}).
% They were originally developed for unsupervised domain adaptation (UDA) or domain generalization. To apply them to our setting, we formulate our target domain as the target domain in UDA and our auxiliary domain as the source domain in UDA. In this way, both the target and auxiliary labels are utilized by adding a target cross-entropy loss on top of their corresponding objective function. 
% As can be seen, although DSN performs the best in both datasets among these methods, their improvements over the existing cervical dysplasia visual inspection methods are quite limited, especially in terms of top-1 accuracy, recall and F1 score. 
% CCSA outperform BrAD, PAC and JCL on the NHS dataset. However, its performance degrades significantly on the Biopsy dataset (see Table~\ref{table:sota_biopsy}).
Such domain adaptation methods mainly focus on solving the domain shift challenge, while ignoring the issue caused by the label inconsistencies that potentially exist in different domains. 
Performing cross-domain knowledge transfer by fetching auxiliary information without selection may introduce undesirable noise to the target domain. Comparatively, we utilize the auxiliary information by first estimating its transferability and then optimizing the classification model jointly with feature alignment in the shared semantic space. 
From the results, we can see that our proposed solution with the adversarial module is more robust and less vulnerable to label inconsistencies across domains. It achieves the best performance in almost all five metrics and outperforms the domain adaptation methods by at least 4.17\% in terms of top-1 accuracy. 

\begin{figure*}[t!]
  \centering
  \includegraphics[width=1.\linewidth]{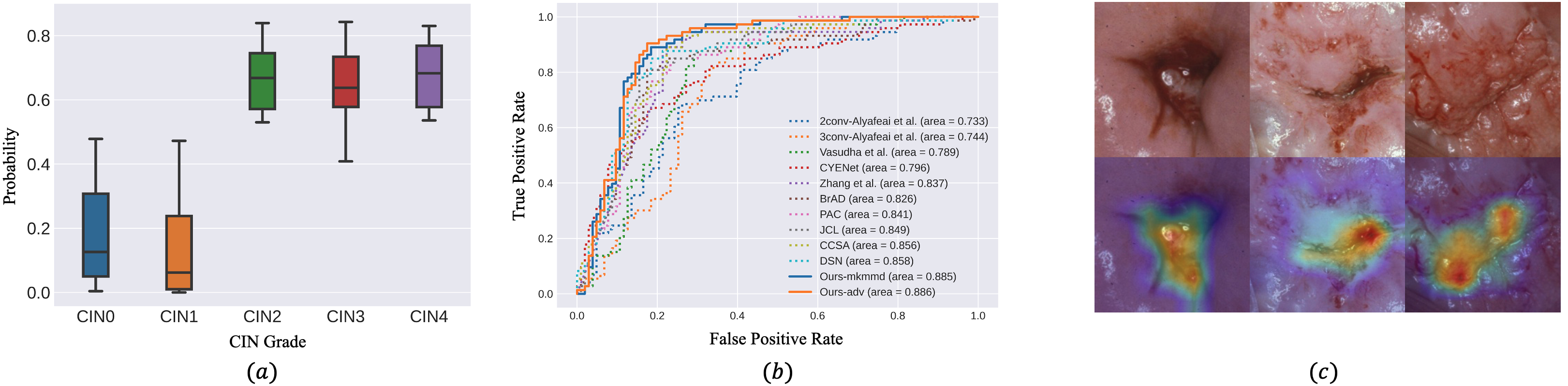}
  \caption{(a) Predicted probability statistics for each CIN grade. (b) ROC curve comparison among methods from both cervical dysplasia visual inspection and domain adaptation. (c) Visualization of model attention based on GradCAM.}
  \label{fig:discussion_3fig}
\end{figure*}

Table~\ref{table:sota_biopsy} reports the performance comparison on the Biopsy dataset. This dataset is highly challenging due to the lower quality and smaller scale compared to the NHS dataset, where only 393 valid records can be utilized for model training. %, and the previous SOTA method~\cite{park2021comparison} obtains a top-1 accuracy of 62.3\% only. Both Park'21 and our method adopt ResNet-50 as the backbone. Our method outperforms Park'21 by 2.45\% in top-1 accuracy with the help of the information learned from the auxiliary domain.
Similarly, our two variants outperform the existing solution in all five metrics, where the adversarial-based early alignment is still better.
Compared with domain adaptation methods, it obtains the best result in four out of the five metrics. 
It outperforms the second best method (\ie DSN) by 2.46\% in top-1 accuracy, 18.89\% in recall, and 9.89\% in F1 score. 
Compared with Table~\ref{table:sota_nhs}, we can see that the performance of domain adaptation methods is less stable on different datasets. One reason might be their heavy dependence on the intra-supervision. Most of them were developed based on the assumption that the given labels are accurately annotated by humans, which is actually not always guaranteed in real-world applications. Our method, on the other hand, utilizes both intra-supervision and inter-supervision simultaneously, thus leading to a more practical and robust solution compared to the previous methods.

\begin{table}[t]
\centering
% \small
% \caption{Different architectures training with auxiliary data.}
\caption{Different architectures training with both domains without filtering.}
\begin{tabular}{c|cc||cc}
\Xhline{1.3pt}
%Architecture & Domain-private Encoders    & Ldi & Accuracy(\%) \\
Architecture & $E_p$    & $\mathcal{L}_{eda}$ & Acc (\%) & Acc Dec. (\%)\\
\hline\hline
$\mathcal{L}_{clf} + \alpha \mathcal{L}_{eda}$         & \cmark &  \cmark   & 82.95 & -       \\
- w/o $\mathcal{L}_{eda}$ & \cmark &   & 82.39 & -0.56    \\
- w/o $E_p$ &  & \cmark  & 82.39 & -0.56    \\
ResNet-50    &    &     & 81.25 &   -1.70      \\

\Xhline{1.3pt}
\end{tabular}
\label{table:abla_arch}
\end{table}

\begin{table}[t]
% \small
% \fontsize{9}{11}\selectfont
\centering

\caption{Ablation study for training strategy.}

%\begin{tabular}{ccc||c|>{\centering\arraybackslash}m{0.35in}}
\begin{tabular}{ccc||c|c}
\Xhline{1.3pt}
\multicolumn{3}{c||}{Training Strategy} &   &      \\
%\cline{1-3}
$\mathcal{L}_{eda}$  & $\mathcal{L}_{psa}$ & $\mathcal{L}_{inter\_clf}$   & \multirow{-2}{*}{Acc (\%)} & \multirow{-2}{*}{Acc Dec.(\%)} \\
\hline\hline
\cmark   & \cmark  & \cmark    & 86.55          & -         \\
     & \cmark  & \cmark    & 83.52        & -3.03     \\
\cmark   &       & \cmark & 84.09         & -2.46      \\
\cmark   &  \cmark &       & 84.09        & -2.46      \\       
\Xhline{1.3pt}
\end{tabular}
\label{table:abla_loss}
\end{table}

\subsection{Ablation Studies}
%In this section, we conduct ablation studies to justify the effectiveness of each component of our proposed framework.

\noindent \textbf{Model Architecture.}
We first set our loss function to $\mathcal{L} = \mathcal{L}_{clf} + \alpha \mathcal{L}_{eda}$ with $\sigma_{clf}=0$ and $\eta_{i}=1$ to remove the impact of the \emph{PSA} module, and evaluate our adversarial-based domain-level alignment in Table~\ref{table:abla_arch}. 
% We first set our loss function to $\mathcal{L} = \mathcal{L}_{ce} + \alpha \mathcal{L}_{ada}$ to remove the impact of the \emph{PDKF} module, and evaluate our \emph{ADA} module design in Table~\ref{table:abla_arch}. 
We compare it with three counterparts given both target data and auxiliary data, including ours without $\mathcal{L}_{eda}$, ours without the domain-private encoders, and single-branch structure without both components. We can see that the top-1 accuracy degrades by 0.56\% if we remove the adversarial loss $\mathcal{L}_{eda}$ from our method or replace the domain-private encoders with a shared encoder. A further degradation of 1.70\% can be observed if both components are removed. Thus, the experimental results verify the effectiveness of our \emph{EDA} module for cervical dysplasia visual inspection.

\noindent \textbf{Training Strategy.} 
Our training strategy consists of multiple loss functions as shown in Eqn.~\ref{loss_all}. Here we examine their effectiveness by removing each of them from $\mathcal{L}$ and report the results in Table~\ref{table:abla_loss}. 
Since the supervision from the target data is necessary for our task, we conduct this ablation study only on $\mathcal{L}_{eda}$, $\mathcal{L}_{psa}$, and $\mathcal{L}_{inter\_clf}$. We observe that, compared to our proposed adversarial-based objective function, removing either one of the individual losses leads to performance degradation ranging from 2.46\% to 3.03\% in top-1 accuracy. 
Among them, $\mathcal{L}_{inter\_clf}$ and $\mathcal{L}_{psa}$ both serve as the bridges for integrating auxiliary knowledge, but from different aspects, thus leading to similar performance decrement.
% Since the intra-supervision for target data is necessary for classification task, we conduct this ablation study only on $\mathcal{L}_{ada}$, $\mathcal{L}_{inter\_sup}$ and $\mathcal{L}_{inter\_con}$. We observed that, compared to our proposed objective function, removing either one of the individual losses leads to performance degradation ranging from 1.70\% to 2.27\% in top-1 accuracy. 
% Among them, $\mathcal{L}_{inter\_sup}$ and $\mathcal{L}_{inter\_con}$ both serves as the bridges for integrating auxiliary knowledge, but from different aspects, thus leading to similar performance decrement.
The results indicate that all our proposed losses are indispensable components of our method, which work collaboratively to complete the task.

\subsection{Discussion}

\noindent \textbf{Statistical Prediction Distribution.}
Our binary setting originally comes from five categories (CIN 0,1,2,3,4) --- CIN0 and CIN1 are regarded as normal, while the rest of them are regarded as abnormal. In Figure~\ref{fig:discussion_3fig}(a), we visualize the statistical prediction distribution of our model for each category, where the Y-axis represents the predicted probability of belonging to the abnormal case. We can observe a distinct margin between the first two levels and the last three. The generally non-overlapped phenomenon between the upper bound of normal classes and the lower bound of abnormal classes reveals a clear decision boundary from our model.

\noindent \textbf{ROC Curve.}
Receiver operating characteristics (ROC) is a probability curve for classification problems at various threshold settings. %is a good measurement for model capability.
In Figure~\ref{fig:discussion_3fig}(b), we present the ROC curves from all methods in Table~\ref{table:sota_nhs} for comparison. The closer the curve to the left-top (\ie the larger the area under the curve (AUC)), the better capability the method has. Our method, shown in the orange line, surpasses all other methods with 0.886 in AUC. %And generally, methods utilizing auxiliary data have better ROC curves, which is consistent with our finding in the previous section. 

\noindent \textbf{Model Attention Visualization.}
Based on GradCAM~\cite{jacobgilpytorchcam,selvaraju2017grad}, we visualize the last convolutional attention map from our framework as shown in Figure~\ref{fig:discussion_3fig}(c). The brighter color in the second row represents the higher-focus area. We can see that % even after the cropping scheme, pixels here are not equally important, 
our model focuses more on the areas with obvious pathological features around the cervix, providing a more reasonable prediction for those patients.

\iffalse
\begin{table}[t]
\centering
\caption{Performance comparison when training with different ratio labeled data.}
\begin{tabular}{c|c|c|c}
\Xhline{1.3pt}
Label Ratio  & ResNet-50 & Ours-KL   & Ours-Adv \\
\hline\hline
5\%   & 70.45  & 81.25    & 81.82    \\
10\%  & 69.89  & 77.84    & 76.70    \\
20\%  & 76.14  & 83.52    & 81.82    \\
50\%  & 82.95  & 83.52    & 84.66    \\  
100\% & 82.95  & 85.80    & \textbf{86.55}    \\
\Xhline{1.3pt}
\end{tabular}
\label{table:abla_tgt_ratio}
\end{table}
\fi

\iffalse
\begin{table}[t]
\centering
\caption{Performance comparison when training with different ratio labeled data.}
\begin{tabular}{c|c|c|c}
\Xhline{1.3pt}
Label Ratio  & ResNet-50 & Ours-mkmmd   & Ours-adv \\
\hline\hline
%5\%   & 70.45  & 73.30    & \textbf{81.82}    \\
10\%  & 69.89  & 75.57    & \textbf{76.70}    \\
20\%  & 76.14  & 80.68    & \textbf{81.82}    \\
50\%  & 81.82  & 82.39    & \textbf{84.66}    \\  
100\% & 82.84  & 85.80    & \textbf{86.55}    \\
\Xhline{1.3pt}
\end{tabular}
\label{table:abla_tgt_ratio}
\end{table}
\fi

\begin{table}[t]
\centering
\caption{Performance comparison when training with different number of target samples.}
\begin{tabular}{c|c|c|c|c}
\Xhline{1.3pt}
Percentage  & 10\% & 20\%   & 50\% & 100\% \\
\hline\hline
ResNet-50   & 69.89 & 76.14 & 81.82 & 82.84   \\
Ours-mkmmd  & 75.57 & 80.68 & 82.39 & 85.80     \\
Ours-adv    & \textbf{76.70}     & \textbf{81.82}  & \textbf{84.66}  & \textbf{86.55}    \\
\Xhline{1.3pt}
\end{tabular}
\label{table:abla_tgt_ratio}
\end{table}

\noindent \textbf{Percentage of target data.} 
We further compare our framework with the baseline model using different percentages of target data and report the results in Table~\ref{table:abla_tgt_ratio}. In each column, we randomly select a certain percentage of target samples for training to study the impact of the target data volume.  
% The results show that our adversarial-based variant surpasses the baseline model by an overall large margin, where 6.81\%, 5.68\%, 2.84\%, and 3.71\% improvement are obtained when the label percentage equals to 10\%, 20\%, 50\%, and 100\%, respectively. 
% It shows that our framework is able to provide complementary information by using auxiliary data, especially when the label is limited on the target domain.
The results show that our adversarial-based approach outperforms the baseline model by a significant margin, with improvements of 6.81\%, 5.68\%, 2.84\%, and 3.71\% achieved when the percentage is set to 10\%, 20\%, 50\%, and 100\%, respectively.
These findings demonstrate that our framework is able to learn complementary and transferable information from the auxiliary data, which is particularly beneficial when the amount of labeled data in the target domain is limited.

% \begin{figure}[t!]
%   \centering
%   \includegraphics[width=.95\linewidth]{fig/discussion_2thres.png}
%   \caption{Performance comparison using different thresholds on the NHS dataset. (a): Different $\sigma_{align}$ with $\sigma_{clf}=0.9$. (b): Different $\sigma_{clf}$ with $\sigma_{align}=0.4$.}
%   \label{fig:discussion_thres}
% \end{figure}

\begin{figure}[t!]
  \centering
  \includegraphics[width=1.\linewidth]{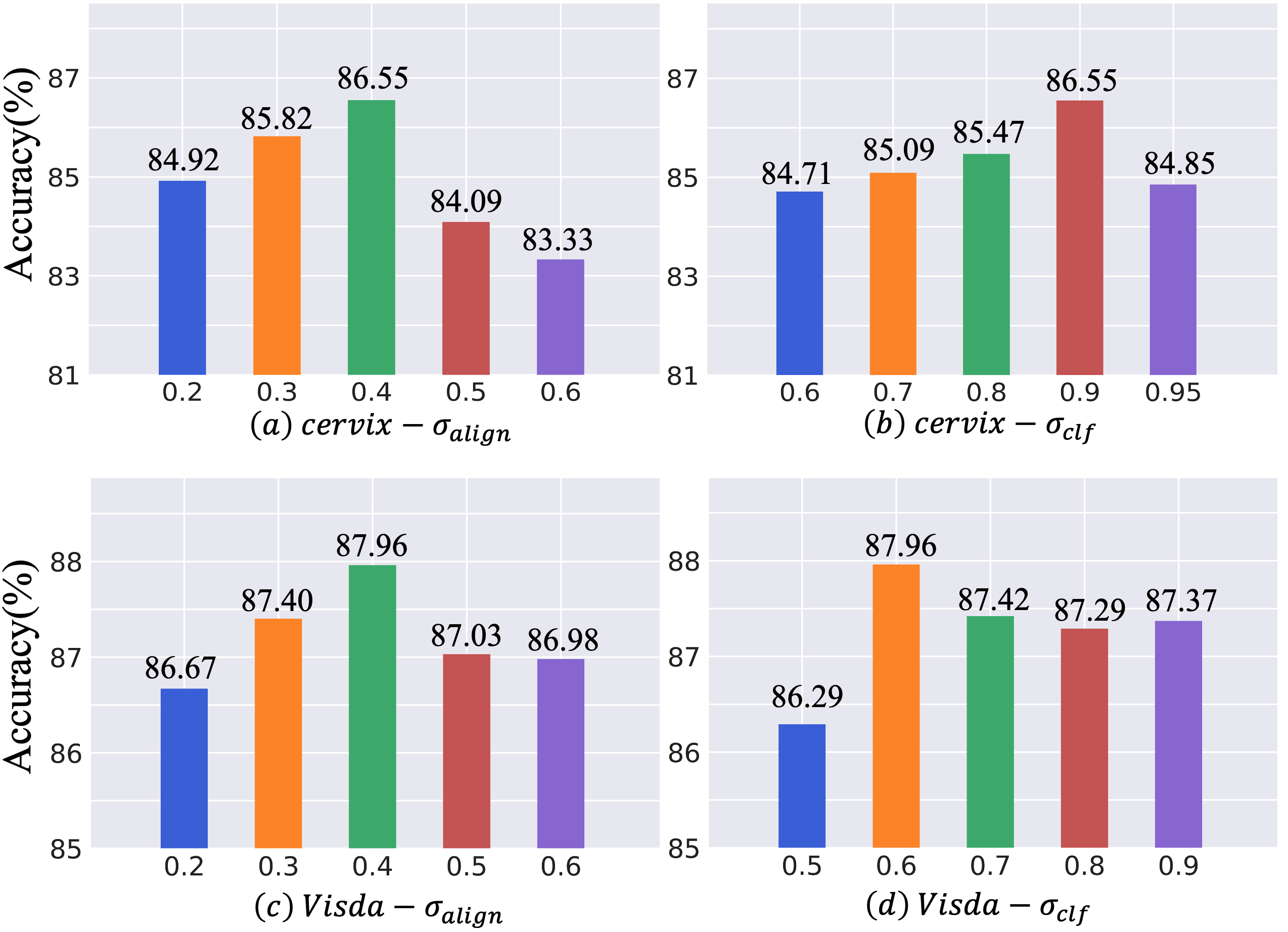}
  \caption[Performance comparison using different thresholds on the NHS and Visda dataset.]{Performance comparison using different thresholds on the NHS and Visda dataset. (a): Different $\sigma_{align}$ with $\sigma_{clf}=0.9$ on NHS dataset. (b): Different $\sigma_{clf}$ with $\sigma_{align}=0.4$ on NHS dataset. (c): Different $\sigma_{align}$ with $\sigma_{clf}=0.6$ on Visda dataset. (d): Different $\sigma_{clf}$ with $\sigma_{align}=0.4$ on Visda dataset.}
  \label{fig:discussion_thres}
\end{figure}

\noindent \textbf{Thresholds.} 
We investigate the impact of two key hyper-parameters, $\sigma_{align}$ and $\sigma_{clf}$, on the NHS dataset and compare the top-1 accuracy of one candidate by holding the other fixed as the best value we found during the experiment.
As shown in Figure~\ref{fig:discussion_thres}(a), the best classification result was obtained with $\sigma_{align}=0.4$. Decreasing $\sigma_{align}$ will result in supervision with noisy labels, while increasing it will result in less information learned and transferred from the auxiliary domain. For $\sigma_{clf}$, which defines the transferability threshold for $\mathcal{L}_{inter\_clf}$, Figure~\ref{fig:discussion_thres}(b) shows that the best classification result was obtained with $\sigma_{clf}=0.9$.
%More than 1\% degradation can be observed if we further decrease this threshold, resulting in supervision based on noisy labels. On the other hand, performance degradation is also observed if we further increase this threshold, resulting in less information learned and transferred from the auxiliary domain.
%For $\sigma_{clf}$, which defines the transferability threshold for $\mathcal{L}_{inter\_clf}$, the best performance was obtained with $\sigma_{clf}=0.9$. 
A similar pattern can be observed that either increasing or decreasing $\sigma_{clf}$ will lead to performance degradation.
However, compared with $\sigma_{clf}$, we observe that decreasing $\sigma_{align}$ has less impact than increasing it. This indicates that contrastive feature alignment is more tolerable with out-of-distribution semantics than direct inter-domain supervision.
In supervised contrastive learning, multiple positive pairs are constructed, both in-domain and cross-domain, making it a more robust solution to the potential inconsistency between cross-domain classification boundaries.

\begin{table}[t]
\centering
\caption{Performance comparison between our method and domain adaptation methods on the Visda-2017 dataset.}
% \fontsize{9}{11}\selectfont
\begin{tabular}{>{\centering\arraybackslash}m{0.85in}|>{\centering\arraybackslash}m{0.85in}>{\centering\arraybackslash}m{0.85in}}
\Xhline{1.3pt}
                     & Top-1 (\%) & Top-5 (\%) \\
\hline\hline
% PAC~\cite{mishra2021surprisingly}           & 66.00   & 92.15   \\
CCSA~\cite{motiian2017unified}          & 77.64   & 97.06   \\
JCL~\cite{Park2020JCL}           & 78.12   & 97.48   \\
BrAD~\cite{harary2022unsupervised}          & 83.89   & 98.29   \\
DSN~\cite{Bousmalis2016DomainSN}           & 84.00   & 97.92   \\
\hline
\textbf{Ours-mkmmd} &  85.59   &   98.32 \\
\textbf{Ours-adv} & \textbf{87.96}    & \textbf{98.69}   \\
\Xhline{1.3pt}
\end{tabular}
\label{table:sota_visda}
\end{table}

\subsection{Results on Visda-2017 Dataset}
To verify the generalization capability of our model, we conduct additional experiments on Visda-2017~\cite{Peng2017VisDATV}. %, which is a benchmark dataset for domain adaptation. 
% Different from the cervix dataset, it is a general image dataset consisting of 152,397 synthetic images and 55,388 real images across 12 classes for training, together with 72,372 real images for validation. 
Different from the cervix dataset, it is a large and general image dataset consisting of synthetic images and real images across 12 classes. 
%the large number of data provides a robust criterion for the measurement of framework capability. 
The potential noise inside the synthetic images due to the artificial generation process is a large obstacle towards good performance. Therefore, both domain shift and label uncertainty challenges are presented in this dataset.
In these experiments, we regard the real images as the target domain and the synthetic images as the auxiliary domain to evaluate our method.
Compared with existing domain adaptation methods~\cite{Bousmalis2016DomainSN,Park2020JCL,motiian2017unified,harary2022unsupervised}, we report the top-1 and top-5 accuracy in Table~\ref{table:sota_visda}. We can see that our framework is able to surpass other domain adaptation methods in this general dataset, obtaining a 3.96\% improvement in top-1 accuracy compared to the second-best solution. 
The results show that our method not only works well in small-scale medical datasets that focus on specific binary classification problem, but also generalizes well in large-scale general image datasets and multi-class classification problems.
We also investigate the impact of the thresholds on Visda-2017. As shown in Figure~\ref{fig:discussion_thres}(c) and (d), the best result is obtained with $\sigma_{clf}=0.6$ and $\sigma_{align}=0.4$. Different from the results on the cervix dataset, a lower $\sigma_{clf}$ is preferable on Visda-2017, possibly due to high cross-domain label consistency.
%and the less sensitivity of decision boundary compared to the medical images. %Thus our method may also capable of measuring the domain correlation in different applications, which is left to the future exploration.
To summarize, the results indicate the potential utilization of our proposed method in applications other than the medical domain, which will be explored as part of our future work.

\section{Conclusion}
Targeted at cervical dysplasia visual inspection, we present a novel prototypical cross-domain knowledge transfer framework to perform robust auxiliary-to-target knowledge transfer. Two key components are introduced in our method, namely the \emph{EDA} module and the \emph{PSA} module. 
The former addresses the domain shift problem by aligning the intermediate representations, while the latter utilizes a prototype-based strategy to learn useful and reliable semantic information from the auxiliary domain.
Experiments on three benchmark cervical image datasets demonstrate the state-of-the-art performance of our proposed approach, with 4.7\% improvement in top-1 accuracy and 0.05 in ROC-AUC. % compared to the existing best solution in cervical dysplasia visual inspection. 
Additional result visualizations and ablation studies are presented to validate our framework design, together with the experiments on Visda-2017 dataset to demonstrate the effectiveness of our method in a more general problem setting.
% Experiments on three benchmark cervical image datasets demonstrate the state-of-the-art performance of our proposed approach, together with additional ablation studies and discussions to validate our framework design. The experiments on Visda-2017 dataset further demonstrate the effectiveness of our method in a more general problem setting.
%In the future, we plan to investigate our method in more real-world applications, and develop more advanced cross-domain knowledge transfer strategies for further improvement.
In the future, we plan to investigate the potential of our method in not only cross-domain but also cross-modal applications with varying label quality.

% \section{Acknowledgments}

%%
%% The acknowledgments section is defined using the "acks" environment
%% (and NOT an unnumbered section). This ensures the proper
%% identification of the section in the article metadata, and the
%% consistent spelling of the heading.
\begin{acks}
This work was supported by Singapore Ministry of Education Academic
Research Fund Tier 1 under MOE's official grant number T1 251RES2029, the National Natural Science Foundation of China No. 62272390, and Zhejiang Gongshang University "Digital+" Disciplinary Construction Management Project (Project Number SZJ2022C005).
\end{acks}

%%
%% The next two lines define the bibliography style to be used, and
%% the bibliography file.
% \clearpage
\bibliographystyle{ACM-Reference-Format}
% \balance
\bibliography{main}

%%% -*-BibTeX-*-
%%% Do NOT edit. File created by BibTeX with style
%%% ACM-Reference-Format-Journals [18-Jan-2012].

\begin{thebibliography}{65}

%%% ====================================================================
%%% NOTE TO THE USER: you can override these defaults by providing
%%% customized versions of any of these macros before the \bibliography
%%% command.  Each of them MUST provide its own final punctuation,
%%% except for \shownote{}, \showDOI{}, and \showURL{}.  The latter two
%%% do not use final punctuation, in order to avoid confusing it with
%%% the Web address.
%%%
%%% To suppress output of a particular field, define its macro to expand
%%% to an empty string, or better, \unskip, like this:
%%%
%%% \newcommand{\showDOI}[1]{\unskip}   % LaTeX syntax
%%%
%%% \def \showDOI #1{\unskip}           % plain TeX syntax
%%%
%%% ====================================================================

\ifx \showCODEN    \undefined \def \showCODEN     #1{\unskip}     \fi
\ifx \showDOI      \undefined \def \showDOI       #1{#1}\fi
\ifx \showISBNx    \undefined \def \showISBNx     #1{\unskip}     \fi
\ifx \showISBNxiii \undefined \def \showISBNxiii  #1{\unskip}     \fi
\ifx \showISSN     \undefined \def \showISSN      #1{\unskip}     \fi
\ifx \showLCCN     \undefined \def \showLCCN      #1{\unskip}     \fi
\ifx \shownote     \undefined \def \shownote      #1{#1}          \fi
\ifx \showarticletitle \undefined \def \showarticletitle #1{#1}   \fi
\ifx \showURL      \undefined \def \showURL       {\relax}        \fi
% The following commands are used for tagged output and should be
% invisible to TeX
\providecommand\bibfield[2]{#2}
\providecommand\bibinfo[2]{#2}
\providecommand\natexlab[1]{#1}
\providecommand\showeprint[2][]{arXiv:#2}

\bibitem[KAG(2017)]%
        {KAGGLE-DATASET}
 \bibinfo{year}{2017}\natexlab{}.
\newblock \bibinfo{title}{Intel\&MobileODT dataset}.
\newblock \bibinfo{howpublished}{\url{https://www.kaggle.com/c/intel-
  mobileodt- cervical - cancer- screening/data}}.
\newblock
\newblock
\shownote{Accessed 2021-02-10}.


\bibitem[imb(2018)]%
        {imbalanced_sampler}
 \bibinfo{year}{2018}\natexlab{}.
\newblock \bibinfo{title}{Imbalanced Sampler}.
\newblock
  \bibinfo{howpublished}{\url{https://github.com/ufoym/imbalanced-dataset-sampler}}.
\newblock
\newblock
\shownote{Accessed 2021-02-10}.


\bibitem[Alyafeai and Ghouti(2020)]%
        {ALYAFEAI2020autopipeline}
\bibfield{author}{\bibinfo{person}{Zaid Alyafeai} {and}
  \bibinfo{person}{Lahouari Ghouti}.} \bibinfo{year}{2020}\natexlab{}.
\newblock \showarticletitle{A fully-automated deep learning pipeline for
  cervical cancer classification}.
\newblock \bibinfo{journal}{\emph{Expert Systems with Applications}}
  \bibinfo{volume}{141} (\bibinfo{year}{2020}), \bibinfo{pages}{112951}.
\newblock


\bibitem[Berthelot et~al\mbox{.}(2021)]%
        {berthelot2021adamatch}
\bibfield{author}{\bibinfo{person}{David Berthelot}, \bibinfo{person}{Rebecca
  Roelofs}, \bibinfo{person}{Kihyuk Sohn}, \bibinfo{person}{Nicholas Carlini},
  {and} \bibinfo{person}{Alex Kurakin}.} \bibinfo{year}{2021}\natexlab{}.
\newblock \showarticletitle{Adamatch: A unified approach to semi-supervised
  learning and domain adaptation}.
\newblock \bibinfo{journal}{\emph{arXiv preprint arXiv:2106.04732}}
  (\bibinfo{year}{2021}).
\newblock


\bibitem[Bousmalis et~al\mbox{.}(2016)]%
        {Bousmalis2016DomainSN}
\bibfield{author}{\bibinfo{person}{Konstantinos Bousmalis},
  \bibinfo{person}{George Trigeorgis}, \bibinfo{person}{Nathan Silberman},
  \bibinfo{person}{Dilip Krishnan}, {and} \bibinfo{person}{Dumitru Erhan}.}
  \bibinfo{year}{2016}\natexlab{}.
\newblock \showarticletitle{Domain separation networks}.
\newblock \bibinfo{journal}{\emph{Advances in neural information processing
  systems}}  \bibinfo{volume}{29} (\bibinfo{year}{2016}).
\newblock


\bibitem[Cao et~al\mbox{.}(2018)]%
        {Cao2018DiDADS}
\bibfield{author}{\bibinfo{person}{Jinming Cao}, \bibinfo{person}{Oren Katzir},
  \bibinfo{person}{Peng Jiang}, \bibinfo{person}{Dani Lischinski},
  \bibinfo{person}{Danny Cohen-Or}, \bibinfo{person}{Changhe Tu}, {and}
  \bibinfo{person}{Yangyan Li}.} \bibinfo{year}{2018}\natexlab{}.
\newblock \showarticletitle{Dida: Disentangled synthesis for domain
  adaptation}.
\newblock \bibinfo{journal}{\emph{arXiv preprint arXiv:1805.08019}}
  (\bibinfo{year}{2018}).
\newblock


\bibitem[Caron et~al\mbox{.}(2021)]%
        {Caron2021EmergingPI}
\bibfield{author}{\bibinfo{person}{Mathilde Caron}, \bibinfo{person}{Hugo
  Touvron}, \bibinfo{person}{Ishan Misra}, \bibinfo{person}{Herv{\'e}
  J{\'e}gou}, \bibinfo{person}{Julien Mairal}, \bibinfo{person}{Piotr
  Bojanowski}, {and} \bibinfo{person}{Armand Joulin}.}
  \bibinfo{year}{2021}\natexlab{}.
\newblock \showarticletitle{Emerging properties in self-supervised vision
  transformers}. In \bibinfo{booktitle}{\emph{Proceedings of the IEEE/CVF
  International Conference on Computer Vision}}. \bibinfo{pages}{9650--9660}.
\newblock


\bibitem[Chae et~al\mbox{.}(2022)]%
        {chae2022attention}
\bibfield{author}{\bibinfo{person}{Jinyeong Chae}, \bibinfo{person}{Ying
  Zhang}, \bibinfo{person}{Roger Zimmermann}, \bibinfo{person}{Dongho Kim},
  {and} \bibinfo{person}{Jihie Kim}.} \bibinfo{year}{2022}\natexlab{}.
\newblock \showarticletitle{An Attention-Based Deep Learning Model with
  Interpretable Patch-Weight Sharing for Diagnosing Cervical Dysplasia}. In
  \bibinfo{booktitle}{\emph{Intelligent Systems and Applications: Proceedings
  of the 2021 Intelligent Systems Conference (IntelliSys) Volume 3}}. Springer,
  \bibinfo{pages}{634--642}.
\newblock


\bibitem[Chandran et~al\mbox{.}(2021)]%
        {chandran2021diagnosis}
\bibfield{author}{\bibinfo{person}{Venkatesan Chandran}, \bibinfo{person}{MG
  Sumithra}, \bibinfo{person}{Alagar Karthick}, \bibinfo{person}{Tony George},
  \bibinfo{person}{M Deivakani}, \bibinfo{person}{Balan Elakkiya},
  \bibinfo{person}{Umashankar Subramaniam}, {and} \bibinfo{person}{S
  Manoharan}.} \bibinfo{year}{2021}\natexlab{}.
\newblock \showarticletitle{Diagnosis of cervical cancer based on ensemble deep
  learning network using colposcopy images}.
\newblock \bibinfo{journal}{\emph{BioMed Research International}}
  \bibinfo{volume}{2021} (\bibinfo{year}{2021}).
\newblock


\bibitem[Chang et~al\mbox{.}(2005)]%
        {Chang2005CombinedRA}
\bibfield{author}{\bibinfo{person}{Sung~K Chang}, \bibinfo{person}{Yvette~N
  Mirabal}, \bibinfo{person}{Edward~Neely Atkinson}, \bibinfo{person}{Dennis~D
  Cox}, \bibinfo{person}{Anais Malpica}, \bibinfo{person}{Michelle Follen},
  {and} \bibinfo{person}{Rebecca~R Richards-Kortum}.}
  \bibinfo{year}{2005}\natexlab{}.
\newblock \showarticletitle{Combined reflectance and fluorescence spectroscopy
  for in vivo detection of cervical pre-cancer}.
\newblock \bibinfo{journal}{\emph{Journal of biomedical optics}}
  \bibinfo{volume}{10}, \bibinfo{number}{2} (\bibinfo{year}{2005}),
  \bibinfo{pages}{024031}.
\newblock


\bibitem[Chen et~al\mbox{.}(2019)]%
        {chen2019progressive}
\bibfield{author}{\bibinfo{person}{Chaoqi Chen}, \bibinfo{person}{Weiping Xie},
  \bibinfo{person}{Wenbing Huang}, \bibinfo{person}{Yu Rong},
  \bibinfo{person}{Xinghao Ding}, \bibinfo{person}{Yue Huang},
  \bibinfo{person}{Tingyang Xu}, {and} \bibinfo{person}{Junzhou Huang}.}
  \bibinfo{year}{2019}\natexlab{}.
\newblock \showarticletitle{Progressive feature alignment for unsupervised
  domain adaptation}. In \bibinfo{booktitle}{\emph{Proceedings of the IEEE/CVF
  conference on computer vision and pattern recognition}}.
  \bibinfo{pages}{627--636}.
\newblock


\bibitem[Chen et~al\mbox{.}(2020b)]%
        {Chen2020SimCLR}
\bibfield{author}{\bibinfo{person}{Ting Chen}, \bibinfo{person}{Simon
  Kornblith}, \bibinfo{person}{Mohammad Norouzi}, {and}
  \bibinfo{person}{Geoffrey Hinton}.} \bibinfo{year}{2020}\natexlab{b}.
\newblock \showarticletitle{A simple framework for contrastive learning of
  visual representations}. In \bibinfo{booktitle}{\emph{International
  conference on machine learning}}. PMLR, \bibinfo{pages}{1597--1607}.
\newblock


\bibitem[Chen et~al\mbox{.}(2020c)]%
        {Chen2020SimCLRv2}
\bibfield{author}{\bibinfo{person}{Ting Chen}, \bibinfo{person}{Simon
  Kornblith}, \bibinfo{person}{Kevin Swersky}, \bibinfo{person}{Mohammad
  Norouzi}, {and} \bibinfo{person}{Geoffrey~E Hinton}.}
  \bibinfo{year}{2020}\natexlab{c}.
\newblock \showarticletitle{Big self-supervised models are strong
  semi-supervised learners}.
\newblock \bibinfo{journal}{\emph{Advances in neural information processing
  systems}}  \bibinfo{volume}{33} (\bibinfo{year}{2020}),
  \bibinfo{pages}{22243--22255}.
\newblock


\bibitem[Chen et~al\mbox{.}(2020a)]%
        {Chen2020MoCov2}
\bibfield{author}{\bibinfo{person}{Xinlei Chen}, \bibinfo{person}{Haoqi Fan},
  \bibinfo{person}{Ross Girshick}, {and} \bibinfo{person}{Kaiming He}.}
  \bibinfo{year}{2020}\natexlab{a}.
\newblock \showarticletitle{Improved baselines with momentum contrastive
  learning}.
\newblock \bibinfo{journal}{\emph{arXiv preprint arXiv:2003.04297}}
  (\bibinfo{year}{2020}).
\newblock


\bibitem[Deng et~al\mbox{.}(2009)]%
        {deng2009imagenet}
\bibfield{author}{\bibinfo{person}{Jia Deng}, \bibinfo{person}{Wei Dong},
  \bibinfo{person}{Richard Socher}, \bibinfo{person}{Li-Jia Li},
  \bibinfo{person}{Kai Li}, {and} \bibinfo{person}{Li Fei-Fei}.}
  \bibinfo{year}{2009}\natexlab{}.
\newblock \showarticletitle{Imagenet: A large-scale hierarchical image
  database}. In \bibinfo{booktitle}{\emph{2009 IEEE conference on computer
  vision and pattern recognition}}. Ieee, \bibinfo{pages}{248--255}.
\newblock


\bibitem[DeSantis et~al\mbox{.}(2007)]%
        {DeSantis2007-SpectroscopicIA}
\bibfield{author}{\bibinfo{person}{Timothy DeSantis}, \bibinfo{person}{Nahida
  Chakhtoura}, \bibinfo{person}{Leo Twiggs}, \bibinfo{person}{Daron Ferris},
  \bibinfo{person}{Manocher Lashgari}, \bibinfo{person}{Lisa Flowers},
  \bibinfo{person}{Mark Faupel}, \bibinfo{person}{Shabbir Bambot},
  \bibinfo{person}{Steven Raab}, {and} \bibinfo{person}{Edward Wilkinson}.}
  \bibinfo{year}{2007}\natexlab{}.
\newblock \showarticletitle{Spectroscopic imaging as a triage test for cervical
  disease: a prospective multicenter clinical trial}.
\newblock \bibinfo{journal}{\emph{Journal of lower genital tract disease}}
  \bibinfo{volume}{11}, \bibinfo{number}{1} (\bibinfo{year}{2007}),
  \bibinfo{pages}{18--24}.
\newblock


\bibitem[Du et~al\mbox{.}(2021)]%
        {du2021cross}
\bibfield{author}{\bibinfo{person}{Zhekai Du}, \bibinfo{person}{Jingjing Li},
  \bibinfo{person}{Hongzu Su}, \bibinfo{person}{Lei Zhu}, {and}
  \bibinfo{person}{Ke Lu}.} \bibinfo{year}{2021}\natexlab{}.
\newblock \showarticletitle{Cross-domain gradient discrepancy minimization for
  unsupervised domain adaptation}. In \bibinfo{booktitle}{\emph{Proceedings of
  the IEEE/CVF conference on computer vision and pattern recognition}}.
  \bibinfo{pages}{3937--3946}.
\newblock


\bibitem[Ganin et~al\mbox{.}(2016)]%
        {ganin2016domain}
\bibfield{author}{\bibinfo{person}{Yaroslav Ganin}, \bibinfo{person}{Evgeniya
  Ustinova}, \bibinfo{person}{Hana Ajakan}, \bibinfo{person}{Pascal Germain},
  \bibinfo{person}{Hugo Larochelle}, \bibinfo{person}{Fran{\c{c}}ois
  Laviolette}, \bibinfo{person}{Mario Marchand}, {and} \bibinfo{person}{Victor
  Lempitsky}.} \bibinfo{year}{2016}\natexlab{}.
\newblock \showarticletitle{Domain-adversarial training of neural networks}.
\newblock \bibinfo{journal}{\emph{The journal of machine learning research}}
  \bibinfo{volume}{17}, \bibinfo{number}{1} (\bibinfo{year}{2016}),
  \bibinfo{pages}{2096--2030}.
\newblock


\bibitem[Ghifary et~al\mbox{.}(2014)]%
        {ghifary2014domain}
\bibfield{author}{\bibinfo{person}{Muhammad Ghifary},
  \bibinfo{person}{W~Bastiaan Kleijn}, {and} \bibinfo{person}{Mengjie Zhang}.}
  \bibinfo{year}{2014}\natexlab{}.
\newblock \showarticletitle{Domain adaptive neural networks for object
  recognition}. In \bibinfo{booktitle}{\emph{PRICAI 2014: Trends in Artificial
  Intelligence: 13th Pacific Rim International Conference on Artificial
  Intelligence, Gold Coast, QLD, Australia, December 1-5, 2014. Proceedings
  13}}. Springer, \bibinfo{pages}{898--904}.
\newblock


\bibitem[Gildenblat and contributors(2021)]%
        {jacobgilpytorchcam}
\bibfield{author}{\bibinfo{person}{Jacob Gildenblat} {and}
  \bibinfo{person}{contributors}.} \bibinfo{year}{2021}\natexlab{}.
\newblock \bibinfo{title}{PyTorch library for CAM methods}.
\newblock
  \bibinfo{howpublished}{\url{https://github.com/jacobgil/pytorch-grad-cam}}.
\newblock


\bibitem[Gotlieb et~al\mbox{.}(2017)]%
        {Gotlieb2017ConstraintBasedVO}
\bibfield{author}{\bibinfo{person}{Arnaud Gotlieb}, \bibinfo{person}{Marine
  Louarn}, \bibinfo{person}{Mari Nygard}, \bibinfo{person}{Tomas Ruiz-Lopez},
  \bibinfo{person}{Sagar Sen}, {and} \bibinfo{person}{Roberta Gori}.}
  \bibinfo{year}{2017}\natexlab{}.
\newblock \showarticletitle{Constraint-based verification of a mobile app game
  designed for nudging people to attend cancer screening}. In
  \bibinfo{booktitle}{\emph{Twenty-Ninth IAAI Conference}}.
\newblock


\bibitem[Group(2003)]%
        {Walker2003ART}
\bibfield{author}{\bibinfo{person}{The ASCUS-LSIL Triage Study~ALTS Group}.}
  \bibinfo{year}{2003}\natexlab{}.
\newblock \showarticletitle{A randomized trial on the management of low-grade
  squamous intraepithelial lesion cytology interpretations}.
\newblock \bibinfo{journal}{\emph{American journal of obstetrics and
  gynecology}} \bibinfo{volume}{188}, \bibinfo{number}{6}
  (\bibinfo{year}{2003}), \bibinfo{pages}{1393--1400}.
\newblock


\bibitem[Harary et~al\mbox{.}(2022)]%
        {harary2022unsupervised}
\bibfield{author}{\bibinfo{person}{Sivan Harary}, \bibinfo{person}{Eli
  Schwartz}, \bibinfo{person}{Assaf Arbelle}, \bibinfo{person}{Peter Staar},
  \bibinfo{person}{Shady Abu-Hussein}, \bibinfo{person}{Elad Amrani},
  \bibinfo{person}{Roei Herzig}, \bibinfo{person}{Amit Alfassy},
  \bibinfo{person}{Raja Giryes}, \bibinfo{person}{Hilde Kuehne},
  {et~al\mbox{.}}} \bibinfo{year}{2022}\natexlab{}.
\newblock \showarticletitle{Unsupervised Domain Generalization by Learning a
  Bridge Across Domains}. In \bibinfo{booktitle}{\emph{Proceedings of the
  IEEE/CVF Conference on Computer Vision and Pattern Recognition}}.
  \bibinfo{pages}{5280--5290}.
\newblock


\bibitem[He et~al\mbox{.}(2020)]%
        {He2020MoCo}
\bibfield{author}{\bibinfo{person}{Kaiming He}, \bibinfo{person}{Haoqi Fan},
  \bibinfo{person}{Yuxin Wu}, \bibinfo{person}{Saining Xie}, {and}
  \bibinfo{person}{Ross Girshick}.} \bibinfo{year}{2020}\natexlab{}.
\newblock \showarticletitle{Momentum contrast for unsupervised visual
  representation learning}. In \bibinfo{booktitle}{\emph{Proceedings of the
  IEEE/CVF conference on computer vision and pattern recognition}}.
  \bibinfo{pages}{9729--9738}.
\newblock


\bibitem[He et~al\mbox{.}(2016)]%
        {He2016resnet}
\bibfield{author}{\bibinfo{person}{Kaiming He}, \bibinfo{person}{Xiangyu
  Zhang}, \bibinfo{person}{Shaoqing Ren}, {and} \bibinfo{person}{Jian Sun}.}
  \bibinfo{year}{2016}\natexlab{}.
\newblock \showarticletitle{Deep residual learning for image recognition}. In
  \bibinfo{booktitle}{\emph{Proceedings of the IEEE conference on computer
  vision and pattern recognition}}. \bibinfo{pages}{770--778}.
\newblock


\bibitem[Herrero et~al\mbox{.}(2000)]%
        {Herrero2000PopulationbasedSO}
\bibfield{author}{\bibinfo{person}{Rolando Herrero}, \bibinfo{person}{Allan
  Hildesheim}, \bibinfo{person}{Concepcion Bratti}, \bibinfo{person}{Mark~E
  Sherman}, \bibinfo{person}{Martha Hutchinson}, \bibinfo{person}{Jorge
  Morales}, \bibinfo{person}{Ileana Balmaceda}, \bibinfo{person}{Mitchell~D
  Greenberg}, \bibinfo{person}{Mario Alfaro}, \bibinfo{person}{Robert~D Burk},
  {et~al\mbox{.}}} \bibinfo{year}{2000}\natexlab{}.
\newblock \showarticletitle{Population-based study of human papillomavirus
  infection and cervical neoplasia in rural Costa Rica}.
\newblock \bibinfo{journal}{\emph{Journal of the National Cancer Institute}}
  \bibinfo{volume}{92}, \bibinfo{number}{6} (\bibinfo{year}{2000}),
  \bibinfo{pages}{464--474}.
\newblock


\bibitem[Hu et~al\mbox{.}(2019)]%
        {Hu2019AnOS}
\bibfield{author}{\bibinfo{person}{Liming Hu}, \bibinfo{person}{David Bell},
  \bibinfo{person}{Sameer Antani}, \bibinfo{person}{Zhiyun Xue},
  \bibinfo{person}{Kai Yu}, \bibinfo{person}{Matthew~P Horning},
  \bibinfo{person}{Noni Gachuhi}, \bibinfo{person}{Benjamin Wilson},
  \bibinfo{person}{Mayoore~S Jaiswal}, \bibinfo{person}{Brian Befano},
  {et~al\mbox{.}}} \bibinfo{year}{2019}\natexlab{}.
\newblock \showarticletitle{An observational study of deep learning and
  automated evaluation of cervical images for cancer screening}.
\newblock \bibinfo{journal}{\emph{JNCI: Journal of the National Cancer
  Institute}} \bibinfo{volume}{111}, \bibinfo{number}{9}
  (\bibinfo{year}{2019}), \bibinfo{pages}{923--932}.
\newblock


\bibitem[Jemal et~al\mbox{.}(2011)]%
        {cancer2002}
\bibfield{author}{\bibinfo{person}{Ahmedin Jemal}, \bibinfo{person}{Freddie
  Bray}, \bibinfo{person}{Melissa~M Center}, \bibinfo{person}{Jacques Ferlay},
  \bibinfo{person}{Elizabeth Ward}, {and} \bibinfo{person}{David Forman}.}
  \bibinfo{year}{2011}\natexlab{}.
\newblock \showarticletitle{Global cancer statistics}.
\newblock \bibinfo{journal}{\emph{CA: a cancer journal for clinicians}}
  \bibinfo{volume}{61}, \bibinfo{number}{2} (\bibinfo{year}{2011}),
  \bibinfo{pages}{69--90}.
\newblock


\bibitem[Jiang et~al\mbox{.}(2020)]%
        {jiang2020implicit}
\bibfield{author}{\bibinfo{person}{Xiang Jiang}, \bibinfo{person}{Qicheng Lao},
  \bibinfo{person}{Stan Matwin}, {and} \bibinfo{person}{Mohammad Havaei}.}
  \bibinfo{year}{2020}\natexlab{}.
\newblock \showarticletitle{Implicit class-conditioned domain alignment for
  unsupervised domain adaptation}. In \bibinfo{booktitle}{\emph{International
  Conference on Machine Learning}}. PMLR, \bibinfo{pages}{4816--4827}.
\newblock


\bibitem[Khosla et~al\mbox{.}(2020)]%
        {Khosla2020SupConLoss}
\bibfield{author}{\bibinfo{person}{Prannay Khosla}, \bibinfo{person}{Piotr
  Teterwak}, \bibinfo{person}{Chen Wang}, \bibinfo{person}{Aaron Sarna},
  \bibinfo{person}{Yonglong Tian}, \bibinfo{person}{Phillip Isola},
  \bibinfo{person}{Aaron Maschinot}, \bibinfo{person}{Ce Liu}, {and}
  \bibinfo{person}{Dilip Krishnan}.} \bibinfo{year}{2020}\natexlab{}.
\newblock \showarticletitle{Supervised contrastive learning}.
\newblock \bibinfo{journal}{\emph{Advances in Neural Information Processing
  Systems}}  \bibinfo{volume}{33} (\bibinfo{year}{2020}),
  \bibinfo{pages}{18661--18673}.
\newblock


\bibitem[Kim et~al\mbox{.}(2020)]%
        {kim2020cross}
\bibfield{author}{\bibinfo{person}{Donghyun Kim}, \bibinfo{person}{Kuniaki
  Saito}, \bibinfo{person}{Tae-Hyun Oh}, \bibinfo{person}{Bryan~A Plummer},
  \bibinfo{person}{Stan Sclaroff}, {and} \bibinfo{person}{Kate Saenko}.}
  \bibinfo{year}{2020}\natexlab{}.
\newblock \showarticletitle{Cross-domain self-supervised learning for domain
  adaptation with few source labels}.
\newblock \bibinfo{journal}{\emph{arXiv preprint arXiv:2003.08264}}
  (\bibinfo{year}{2020}).
\newblock


\bibitem[Li et~al\mbox{.}(2019)]%
        {Li2019ASF}
\bibfield{author}{\bibinfo{person}{Chen Li}, \bibinfo{person}{Dan Xue},
  \bibinfo{person}{Zhijie Hu}, \bibinfo{person}{Hao Chen},
  \bibinfo{person}{Yudong Yao}, \bibinfo{person}{Yong Zhang},
  \bibinfo{person}{Mo Li}, \bibinfo{person}{Qian Wang}, {and}
  \bibinfo{person}{Ning Xu}.} \bibinfo{year}{2019}\natexlab{}.
\newblock \showarticletitle{A survey for breast histopathology image analysis
  using classical and deep neural networks}. In
  \bibinfo{booktitle}{\emph{International Conference on Information
  Technologies in Biomedicine}}. Springer, \bibinfo{pages}{222--233}.
\newblock


\bibitem[Li et~al\mbox{.}(2021)]%
        {li2021systematic}
\bibfield{author}{\bibinfo{person}{Johann Li}, \bibinfo{person}{Guangming Zhu},
  \bibinfo{person}{Cong Hua}, \bibinfo{person}{Mingtao Feng},
  \bibinfo{person}{Ping Li}, \bibinfo{person}{Xiaoyuan Lu},
  \bibinfo{person}{Juan Song}, \bibinfo{person}{Peiyi Shen},
  \bibinfo{person}{Xu Xu}, \bibinfo{person}{Lin Mei}, {et~al\mbox{.}}}
  \bibinfo{year}{2021}\natexlab{}.
\newblock \showarticletitle{A Systematic Collection of Medical Image Datasets
  for Deep Learning}.
\newblock \bibinfo{journal}{\emph{arXiv preprint arXiv:2106.12864}}
  (\bibinfo{year}{2021}).
\newblock


\bibitem[Liu and Tuzel(2016)]%
        {liu2016coupled}
\bibfield{author}{\bibinfo{person}{Ming-Yu Liu} {and} \bibinfo{person}{Oncel
  Tuzel}.} \bibinfo{year}{2016}\natexlab{}.
\newblock \showarticletitle{Coupled generative adversarial networks}.
\newblock \bibinfo{journal}{\emph{Advances in neural information processing
  systems}}  \bibinfo{volume}{29} (\bibinfo{year}{2016}).
\newblock


\bibitem[Liu et~al\mbox{.}(2021)]%
        {CVPR2021DualConsecutiveNetwork}
\bibfield{author}{\bibinfo{person}{Zhenguang Liu}, \bibinfo{person}{Haoming
  Chen}, \bibinfo{person}{Runyang Feng}, \bibinfo{person}{Shuang Wu},
  \bibinfo{person}{Shouling Ji}, \bibinfo{person}{Bailin Yang}, {and}
  \bibinfo{person}{Xun Wang}.} \bibinfo{year}{2021}\natexlab{}.
\newblock \showarticletitle{Deep Dual Consecutive Network for Human Pose
  Estimation}. In \bibinfo{booktitle}{\emph{CVPR}}. \bibinfo{pages}{525--534}.
\newblock
\urldef\tempurl%
\url{https://doi.org/10.1109/CVPR46437.2021.00059}
\showDOI{\tempurl}


\bibitem[Liu et~al\mbox{.}(2019)]%
        {CVPR2019MotionPrediction}
\bibfield{author}{\bibinfo{person}{Zhenguang Liu}, \bibinfo{person}{Shuang Wu},
  \bibinfo{person}{Shuyuan Jin}, \bibinfo{person}{Qi Liu},
  \bibinfo{person}{Shijian Lu}, \bibinfo{person}{Roger Zimmermann}, {and}
  \bibinfo{person}{Li Cheng}.} \bibinfo{year}{2019}\natexlab{}.
\newblock \showarticletitle{Towards Natural and Accurate Future Motion
  Prediction of Humans and Animals}. In \bibinfo{booktitle}{\emph{CVPR}}.
  \bibinfo{pages}{10004--10012}.
\newblock
\urldef\tempurl%
\url{https://doi.org/10.1109/CVPR.2019.01024}
\showDOI{\tempurl}


\bibitem[Long et~al\mbox{.}(2015)]%
        {Long2015LearningTF}
\bibfield{author}{\bibinfo{person}{Mingsheng Long}, \bibinfo{person}{Yue Cao},
  \bibinfo{person}{Jianmin Wang}, {and} \bibinfo{person}{Michael Jordan}.}
  \bibinfo{year}{2015}\natexlab{}.
\newblock \showarticletitle{Learning transferable features with deep adaptation
  networks}. In \bibinfo{booktitle}{\emph{International conference on machine
  learning}}. PMLR, \bibinfo{pages}{97--105}.
\newblock


\bibitem[Long et~al\mbox{.}(2017)]%
        {long2017deep}
\bibfield{author}{\bibinfo{person}{Mingsheng Long}, \bibinfo{person}{Han Zhu},
  \bibinfo{person}{Jianmin Wang}, {and} \bibinfo{person}{Michael~I Jordan}.}
  \bibinfo{year}{2017}\natexlab{}.
\newblock \showarticletitle{Deep transfer learning with joint adaptation
  networks}. In \bibinfo{booktitle}{\emph{International conference on machine
  learning}}. PMLR, \bibinfo{pages}{2208--2217}.
\newblock


\bibitem[Mishra et~al\mbox{.}(2021)]%
        {mishra2021surprisingly}
\bibfield{author}{\bibinfo{person}{Samarth Mishra}, \bibinfo{person}{Kate
  Saenko}, {and} \bibinfo{person}{Venkatesh Saligrama}.}
  \bibinfo{year}{2021}\natexlab{}.
\newblock \showarticletitle{Surprisingly simple semi-supervised domain
  adaptation with pretraining and consistency}.
\newblock \bibinfo{journal}{\emph{arXiv preprint arXiv:2101.12727}}
  (\bibinfo{year}{2021}).
\newblock


\bibitem[Motiian et~al\mbox{.}(2017)]%
        {motiian2017unified}
\bibfield{author}{\bibinfo{person}{Saeid Motiian}, \bibinfo{person}{Marco
  Piccirilli}, \bibinfo{person}{Donald~A Adjeroh}, {and}
  \bibinfo{person}{Gianfranco Doretto}.} \bibinfo{year}{2017}\natexlab{}.
\newblock \showarticletitle{Unified deep supervised domain adaptation and
  generalization}. In \bibinfo{booktitle}{\emph{Proceedings of the IEEE
  international conference on computer vision}}. \bibinfo{pages}{5715--5725}.
\newblock


\bibitem[Ou et~al\mbox{.}(2020)]%
        {ou2020semi}
\bibfield{author}{\bibinfo{person}{Yanglan Ou}, \bibinfo{person}{Yuan Xue},
  \bibinfo{person}{Ye Yuan}, \bibinfo{person}{Tao Xu}, \bibinfo{person}{Vincent
  Pisztora}, \bibinfo{person}{Jia Li}, {and} \bibinfo{person}{Xiaolei Huang}.}
  \bibinfo{year}{2020}\natexlab{}.
\newblock \showarticletitle{Semi-supervised cervical dysplasia classification
  with learnable graph convolutional network}. In
  \bibinfo{booktitle}{\emph{2020 IEEE 17th International Symposium on
  Biomedical Imaging (ISBI)}}. IEEE, \bibinfo{pages}{1720--1724}.
\newblock


\bibitem[Park et~al\mbox{.}(2020)]%
        {Park2020JCL}
\bibfield{author}{\bibinfo{person}{Changhwa Park}, \bibinfo{person}{Jonghyun
  Lee}, \bibinfo{person}{Jaeyoon Yoo}, \bibinfo{person}{Minhoe Hur}, {and}
  \bibinfo{person}{Sungroh Yoon}.} \bibinfo{year}{2020}\natexlab{}.
\newblock \showarticletitle{Joint contrastive learning for unsupervised domain
  adaptation}.
\newblock \bibinfo{journal}{\emph{arXiv preprint arXiv:2006.10297}}
  (\bibinfo{year}{2020}).
\newblock


\bibitem[Park et~al\mbox{.}(2021)]%
        {park2021comparison}
\bibfield{author}{\bibinfo{person}{Ye~Rang Park}, \bibinfo{person}{Young~Jae
  Kim}, \bibinfo{person}{Woong Ju}, \bibinfo{person}{Kyehyun Nam},
  \bibinfo{person}{Soonyung Kim}, {and} \bibinfo{person}{Kwang~Gi Kim}.}
  \bibinfo{year}{2021}\natexlab{}.
\newblock \showarticletitle{Comparison of machine and deep learning for the
  classification of cervical cancer based on cervicography images}.
\newblock \bibinfo{journal}{\emph{Scientific Reports}} \bibinfo{volume}{11},
  \bibinfo{number}{1} (\bibinfo{year}{2021}), \bibinfo{pages}{1--11}.
\newblock


\bibitem[Paul(2017)]%
        {KAGGLE-LABEL}
\bibfield{author}{\bibinfo{person}{Paul}.} \bibinfo{year}{2017}\natexlab{}.
\newblock \bibinfo{title}{Kaggle bounding box labels}.
\newblock
  \bibinfo{howpublished}{\url{https://www.kaggle.com/c/intel-mobileodt-cervical-cancer-screening/discussion/31565}}.
\newblock
\newblock
\shownote{Accessed 2021-02-10}.


\bibitem[Peng et~al\mbox{.}(2019)]%
        {peng2019domain}
\bibfield{author}{\bibinfo{person}{Xingchao Peng}, \bibinfo{person}{Zijun
  Huang}, \bibinfo{person}{Ximeng Sun}, {and} \bibinfo{person}{Kate Saenko}.}
  \bibinfo{year}{2019}\natexlab{}.
\newblock \showarticletitle{Domain agnostic learning with disentangled
  representations}. In \bibinfo{booktitle}{\emph{International Conference on
  Machine Learning}}. PMLR, \bibinfo{pages}{5102--5112}.
\newblock


\bibitem[Peng and Saenko(2018)]%
        {peng2018synthetic}
\bibfield{author}{\bibinfo{person}{Xingchao Peng} {and} \bibinfo{person}{Kate
  Saenko}.} \bibinfo{year}{2018}\natexlab{}.
\newblock \showarticletitle{Synthetic to real adaptation with generative
  correlation alignment networks}. In \bibinfo{booktitle}{\emph{2018 IEEE
  Winter Conference on Applications of Computer Vision (WACV)}}. IEEE,
  \bibinfo{pages}{1982--1991}.
\newblock


\bibitem[Peng et~al\mbox{.}(2017)]%
        {Peng2017VisDATV}
\bibfield{author}{\bibinfo{person}{Xingchao Peng}, \bibinfo{person}{Ben Usman},
  \bibinfo{person}{Neela Kaushik}, \bibinfo{person}{Judy Hoffman},
  \bibinfo{person}{Dequan Wang}, {and} \bibinfo{person}{Kate Saenko}.}
  \bibinfo{year}{2017}\natexlab{}.
\newblock \showarticletitle{VisDA: The Visual Domain Adaptation Challenge}.
\newblock \bibinfo{journal}{\emph{ArXiv}}  \bibinfo{volume}{abs/1710.06924}
  (\bibinfo{year}{2017}).
\newblock


\bibitem[Raghu et~al\mbox{.}(2019)]%
        {raghu2019transfusion}
\bibfield{author}{\bibinfo{person}{Maithra Raghu}, \bibinfo{person}{Chiyuan
  Zhang}, \bibinfo{person}{Jon Kleinberg}, {and} \bibinfo{person}{Samy
  Bengio}.} \bibinfo{year}{2019}\natexlab{}.
\newblock \showarticletitle{Transfusion: Understanding transfer learning for
  medical imaging}.
\newblock \bibinfo{journal}{\emph{Advances in neural information processing
  systems}}  \bibinfo{volume}{32} (\bibinfo{year}{2019}).
\newblock


\bibitem[Ren et~al\mbox{.}(2015)]%
        {Ren2015FasterRCNN}
\bibfield{author}{\bibinfo{person}{Shaoqing Ren}, \bibinfo{person}{Kaiming He},
  \bibinfo{person}{Ross Girshick}, {and} \bibinfo{person}{Jian Sun}.}
  \bibinfo{year}{2015}\natexlab{}.
\newblock \showarticletitle{Faster r-cnn: Towards real-time object detection
  with region proposal networks}.
\newblock \bibinfo{journal}{\emph{Advances in neural information processing
  systems}}  \bibinfo{volume}{28} (\bibinfo{year}{2015}).
\newblock


\bibitem[Saini et~al\mbox{.}(2020)]%
        {saini2020colponet}
\bibfield{author}{\bibinfo{person}{Sumindar~Kaur Saini},
  \bibinfo{person}{Vasudha Bansal}, \bibinfo{person}{Ravinder Kaur}, {and}
  \bibinfo{person}{Mamta Juneja}.} \bibinfo{year}{2020}\natexlab{}.
\newblock \showarticletitle{ColpoNet for automated cervical cancer screening
  using colposcopy images}.
\newblock \bibinfo{journal}{\emph{Machine Vision and Applications}}
  \bibinfo{volume}{31} (\bibinfo{year}{2020}), \bibinfo{pages}{1--15}.
\newblock


\bibitem[Saito et~al\mbox{.}(2019)]%
        {saito2019semi}
\bibfield{author}{\bibinfo{person}{Kuniaki Saito}, \bibinfo{person}{Donghyun
  Kim}, \bibinfo{person}{Stan Sclaroff}, \bibinfo{person}{Trevor Darrell},
  {and} \bibinfo{person}{Kate Saenko}.} \bibinfo{year}{2019}\natexlab{}.
\newblock \showarticletitle{Semi-supervised domain adaptation via minimax
  entropy}. In \bibinfo{booktitle}{\emph{Proceedings of the IEEE/CVF
  international conference on computer vision}}. \bibinfo{pages}{8050--8058}.
\newblock


\bibitem[Saito et~al\mbox{.}(2017)]%
        {saito2017asymmetric}
\bibfield{author}{\bibinfo{person}{Kuniaki Saito}, \bibinfo{person}{Yoshitaka
  Ushiku}, {and} \bibinfo{person}{Tatsuya Harada}.}
  \bibinfo{year}{2017}\natexlab{}.
\newblock \showarticletitle{Asymmetric tri-training for unsupervised domain
  adaptation}. In \bibinfo{booktitle}{\emph{International Conference on Machine
  Learning}}. PMLR, \bibinfo{pages}{2988--2997}.
\newblock


\bibitem[Saito et~al\mbox{.}(2018)]%
        {saito2018maximum}
\bibfield{author}{\bibinfo{person}{Kuniaki Saito}, \bibinfo{person}{Kohei
  Watanabe}, \bibinfo{person}{Yoshitaka Ushiku}, {and} \bibinfo{person}{Tatsuya
  Harada}.} \bibinfo{year}{2018}\natexlab{}.
\newblock \showarticletitle{Maximum classifier discrepancy for unsupervised
  domain adaptation}. In \bibinfo{booktitle}{\emph{Proceedings of the IEEE
  conference on computer vision and pattern recognition}}.
  \bibinfo{pages}{3723--3732}.
\newblock


\bibitem[Selvaraju et~al\mbox{.}(2017)]%
        {selvaraju2017grad}
\bibfield{author}{\bibinfo{person}{Ramprasaath~R Selvaraju},
  \bibinfo{person}{Michael Cogswell}, \bibinfo{person}{Abhishek Das},
  \bibinfo{person}{Ramakrishna Vedantam}, \bibinfo{person}{Devi Parikh}, {and}
  \bibinfo{person}{Dhruv Batra}.} \bibinfo{year}{2017}\natexlab{}.
\newblock \showarticletitle{Grad-cam: Visual explanations from deep networks
  via gradient-based localization}. In \bibinfo{booktitle}{\emph{Proceedings of
  the IEEE international conference on computer vision}}.
  \bibinfo{pages}{618--626}.
\newblock


\bibitem[Song et~al\mbox{.}(2014)]%
        {MultimodalEC-Song2015}
\bibfield{author}{\bibinfo{person}{Dezhao Song}, \bibinfo{person}{Edward Kim},
  \bibinfo{person}{Xiaolei Huang}, \bibinfo{person}{Joseph Patruno},
  \bibinfo{person}{H{\'e}ctor Mu{\~n}oz-Avila}, \bibinfo{person}{Jeff Heflin},
  \bibinfo{person}{L~Rodney Long}, {and} \bibinfo{person}{Sameer Antani}.}
  \bibinfo{year}{2014}\natexlab{}.
\newblock \showarticletitle{Multimodal entity coreference for cervical
  dysplasia diagnosis}.
\newblock \bibinfo{journal}{\emph{IEEE transactions on medical imaging}}
  \bibinfo{volume}{34}, \bibinfo{number}{1} (\bibinfo{year}{2014}),
  \bibinfo{pages}{229--245}.
\newblock


\bibitem[Stoler et~al\mbox{.}(2015)]%
        {biopsy}
\bibfield{author}{\bibinfo{person}{Mark~H Stoler}, \bibinfo{person}{Brigitte~M
  Ronnett}, \bibinfo{person}{Nancy~E Joste}, \bibinfo{person}{William~C Hunt},
  \bibinfo{person}{Jack Cuzick}, {and} \bibinfo{person}{Cosette~M Wheeler}.}
  \bibinfo{year}{2015}\natexlab{}.
\newblock \showarticletitle{The interpretive variability of cervical biopsies
  and its relationship to HPV status}.
\newblock \bibinfo{journal}{\emph{The American journal of surgical pathology}}
  \bibinfo{volume}{39}, \bibinfo{number}{6} (\bibinfo{year}{2015}),
  \bibinfo{pages}{729}.
\newblock


\bibitem[Sun and Saenko(2016)]%
        {sun2016deep}
\bibfield{author}{\bibinfo{person}{Baochen Sun} {and} \bibinfo{person}{Kate
  Saenko}.} \bibinfo{year}{2016}\natexlab{}.
\newblock \showarticletitle{Deep coral: Correlation alignment for deep domain
  adaptation}. In \bibinfo{booktitle}{\emph{Computer Vision--ECCV 2016
  Workshops: Amsterdam, The Netherlands, October 8-10 and 15-16, 2016,
  Proceedings, Part III 14}}. Springer, \bibinfo{pages}{443--450}.
\newblock


\bibitem[Vasudha and Juneja(2018)]%
        {Vasudha2018CervixCC}
\bibfield{author}{\bibinfo{person}{Ajay~Mittal Vasudha} {and}
  \bibinfo{person}{Mamta Juneja}.} \bibinfo{year}{2018}\natexlab{}.
\newblock \showarticletitle{Cervix cancer classification using colposcopy
  images by deep learning method}.
\newblock \bibinfo{journal}{\emph{Int J Eng Technol Sci Res}}
  \bibinfo{volume}{5} (\bibinfo{year}{2018}), \bibinfo{pages}{426--432}.
\newblock


\bibitem[WHO(2022)]%
        {WHO-cancer}
\bibfield{author}{\bibinfo{person}{World Health~Organization WHO}.}
  \bibinfo{year}{2022}\natexlab{}.
\newblock \bibinfo{title}{WHO-cervical-cancer}.
\newblock
  \bibinfo{howpublished}{\url{https://www.who.int/health-topics/cervical-cancer}}.
\newblock
\newblock
\shownote{Accessed 2021-07-15}.


\bibitem[Xu et~al\mbox{.}(2016)]%
        {Xu2016MultimodalDL}
\bibfield{author}{\bibinfo{person}{Tao Xu}, \bibinfo{person}{Han Zhang},
  \bibinfo{person}{Xiaolei Huang}, \bibinfo{person}{Shaoting Zhang}, {and}
  \bibinfo{person}{Dimitris~N Metaxas}.} \bibinfo{year}{2016}\natexlab{}.
\newblock \showarticletitle{Multimodal deep learning for cervical dysplasia
  diagnosis}. In \bibinfo{booktitle}{\emph{International conference on medical
  image computing and computer-assisted intervention}}. Springer,
  \bibinfo{pages}{115--123}.
\newblock


\bibitem[Yang et~al\mbox{.}(2020)]%
        {yang2020label}
\bibfield{author}{\bibinfo{person}{Jinyu Yang}, \bibinfo{person}{Weizhi An},
  \bibinfo{person}{Sheng Wang}, \bibinfo{person}{Xinliang Zhu},
  \bibinfo{person}{Chaochao Yan}, {and} \bibinfo{person}{Junzhou Huang}.}
  \bibinfo{year}{2020}\natexlab{}.
\newblock \showarticletitle{Label-driven reconstruction for domain adaptation
  in semantic segmentation}. In \bibinfo{booktitle}{\emph{Computer Vision--ECCV
  2020: 16th European Conference, Glasgow, UK, August 23--28, 2020,
  Proceedings, Part XXVII 16}}. Springer, \bibinfo{pages}{480--498}.
\newblock


\bibitem[Yin et~al\mbox{.}(2021)]%
        {yin2021enhanced}
\bibfield{author}{\bibinfo{person}{Yifang Yin}, \bibinfo{person}{Harsh
  Shrivastava}, \bibinfo{person}{Ying Zhang}, \bibinfo{person}{Zhenguang Liu},
  \bibinfo{person}{Rajiv~Ratn Shah}, {and} \bibinfo{person}{Roger Zimmermann}.}
  \bibinfo{year}{2021}\natexlab{}.
\newblock \showarticletitle{Enhanced audio tagging via multi-to single-modal
  teacher-student mutual learning}. In \bibinfo{booktitle}{\emph{Proceedings of
  the AAAI conference on artificial intelligence}}, Vol.~\bibinfo{volume}{35}.
  \bibinfo{pages}{10709--10717}.
\newblock


\bibitem[Zellinger et~al\mbox{.}(2017)]%
        {zellinger2017central}
\bibfield{author}{\bibinfo{person}{Werner Zellinger}, \bibinfo{person}{Thomas
  Grubinger}, \bibinfo{person}{Edwin Lughofer}, \bibinfo{person}{Thomas
  Natschl{\"a}ger}, {and} \bibinfo{person}{Susanne Saminger-Platz}.}
  \bibinfo{year}{2017}\natexlab{}.
\newblock \showarticletitle{Central moment discrepancy (cmd) for
  domain-invariant representation learning}.
\newblock \bibinfo{journal}{\emph{arXiv preprint arXiv:1702.08811}}
  (\bibinfo{year}{2017}).
\newblock


\bibitem[Zhang et~al\mbox{.}(2015)]%
        {zhang2015deep}
\bibfield{author}{\bibinfo{person}{Xu Zhang}, \bibinfo{person}{Felix~Xinnan
  Yu}, \bibinfo{person}{Shih-Fu Chang}, {and} \bibinfo{person}{Shengjin Wang}.}
  \bibinfo{year}{2015}\natexlab{}.
\newblock \showarticletitle{Deep transfer network: Unsupervised domain
  adaptation}.
\newblock \bibinfo{journal}{\emph{arXiv preprint arXiv:1503.00591}}
  (\bibinfo{year}{2015}).
\newblock


\bibitem[Zhang et~al\mbox{.}(2021)]%
        {zhanga2021spatial}
\bibfield{author}{\bibinfo{person}{Ying Zhang}, \bibinfo{person}{Yifang Yin},
  \bibinfo{person}{Zhenguang Liu}, {and} \bibinfo{person}{Roger Zimmermann}.}
  \bibinfo{year}{2021}\natexlab{}.
\newblock \showarticletitle{A Spatial Regulated Patch-Wise Approach for
  Cervical Dysplasia Diagnosis}. In \bibinfo{booktitle}{\emph{Proceedings of
  the AAAI Conference on Artificial Intelligence}}, Vol.~\bibinfo{volume}{35}.
  \bibinfo{pages}{733--740}.
\newblock


\end{thebibliography}
% \clearpage
%%
%% If your work has an appendix, this is the place to put it.

\appendix
\section{Appendix}

% \section{SUPPLEMANTARY MATERIAL}

\subsection{Cervix dataset}
\label{app:cervix_dataset}

\begin{figure}[h!]
  \centering
  \includegraphics[width=.7\linewidth]{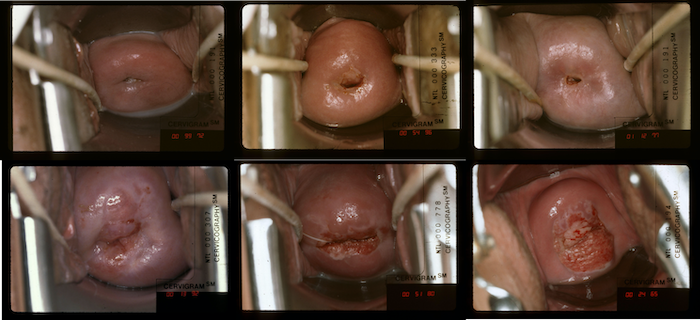}
  \caption{The original cervical images from the NHS dataset. Images in the first row are normal cases while images in the second row are abnormal cases.}
  \label{exp:fig:cervix_ori}
\end{figure}

We utilize a totally of 17,002 cervical images from the Natural History Study of HPV and Cervical Neoplasia (NHS)~\cite{Herrero2000PopulationbasedSO}, ASCUS-LSIL Triage Study (ALTS)~\cite{Walker2003ART} and Biopsy Study (Biopsy)~\cite{biopsy} in this paper. They are three separate clinical studies by the National Cancer Institute (NCI) during previous decades. 
\begin{itemize}
\item NHS is a longitudinal study in Costa Rica started in June 1993, which focuses on studying the role of human papillomavirus infection in the etiology of high-grade cervical neoplasia and evaluating new cervical cancer screening technologies. During 7 years, 10,000 women were enrolled and two cervigrams were taken at each clinic visit as shown in Fig.~\ref{exp:fig:cervix_ori}. It consists of high-resolution cervical images with the shape of around $2400\times1600\times3$. 
\item ALTS was designed to evaluate 3 alternative methods (immediate colposcopy, repeat PAP tests and testing for HPV) for managing atypical squamous cells of undetermined significance (ASCUS) and low-grade squamous intraepithelial lesions (LSIL). It is a randomized clinical trial started in November 1996, where women age 18+ with ASCUS (n=3488) or LSIL (n=1572) cytology were enrolled at 4 colposcopy clinics in the United States. Similarly, two cervigrams with shapes similar to the NHS dataset were taken during each visit in this study.
\item Biopsy was a cross-sectional study designed to understand cervical disease on the lesion level and to establish criteria for conducting cervical biopsies. Out of 2,270 women referred for colposcopy, 690 eligible women consented to participate in the study.
\end{itemize}
During these projects, each patient may have participated in multiple screening sessions, where two photographs of the cervix (cervigrams) were taken during each recruitment and clinic visit as shown in Figure~\ref{exp:fig:cervix_ori}. 

The cervical intraepithelial neoplasia (CIN) level normally serves as the criterion to judge the severity of cervical cancer. In our dataset, cervical images are labeled from CIN0 to CIN4, where histologic CIN2 or worse (CIN2+: CIN2, CIN3, CIN4) indicates the cancer precursor or cancer. To construct an appropriate dataset for the model training, which aims at alerting potential patients for further medical examination, we model this problem as a binary classification problem. Cases with CIN2+ are regarded as abnormal cases, while others are regarded as normal cases. Also, abnormal cases whose screening dates surpass one year are discarded due to the possible noise introduced by these samples. In this way, we have 885 images for the NHS dataset, 15,724 images for the ALTS dataset, and 393 images for the Biopsy dataset. 
The positive and negative ratios are 354:531 for the NHS dataset, 1961:13763 for the ALTS dataset and 151:242 for the Biopsy dataset. Two target datasets (NHS and Biopsy) in our case are not largely imbalanced, while the auxiliary dataset is. Thus, we apply a balance sampler~\cite{imbalanced_sampler} to handle the imbalance problem in the auxiliary dataset. For each epoch, we randomly select a balanced subset of the auxiliary samples, which has the same number of images as the target training dataset.
A train-test ratio of 4:1 is further employed on these datasets following \cite{zhanga2021spatial}. 
Specifically, we split the samples based on the session ID (i.e. patient). During each session, two photographs of a patient was taken. Both pictures from the same session (i.e. same patient) will be assigned to either the training set or the testing set.
%Following \cite{zhang2021spatial}, we split out a validation set from the NHS dataset for performance evaluation and assign binary labels for these samples according to their original labels.
The accessibility of these datasets is based on request and constrained agreement.

\begin{figure}[ht!]
  \centering
  \includegraphics[width=.7\linewidth]{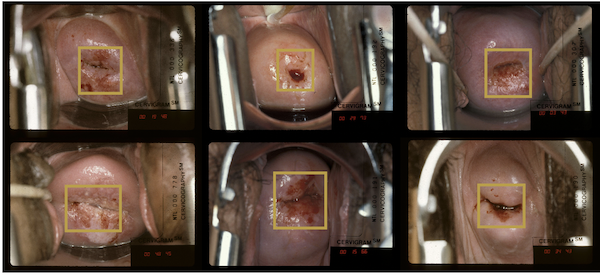}
  \caption{Detection result from our detector with confidence threshold 0.8.}
  \label{fig:detect_result}
\end{figure}

\subsection{Implementation Details}
%\label{app:training}
%\subsection{Data Preprocessing}
Our implementation is based on PyTorch 1.10 and Ubuntu 20.04.
During preprocessing, due to the high resolution of cervical images of 2,400$\times$1,600, we train a cervix detector following \cite{ALYAFEAI2020autopipeline} and use the detector to focus on the important area. 
We crop out the cervix area from the original cervical images, where medical instruments and background still exist, so as to alleviate the problem of information loss when we resize them into $224\times224$ for the better utilization of the ImageNet pre-trained model. To achieve that, we utilize the Intel\&MobileODT dataset~\cite{KAGGLE-DATASET,KAGGLE-LABEL} to train a cervix detector, which is adopted to detect the cervix areas (RoI) from our original cervical images. These areas are later cropped out so as to build a fresh dataset with smaller-shape cervical images (cropped cervical images). The confidence threshold of detection is set to $0.8$ so as to discard invalid or unreliable cropping results that may interfere with model training. The detection results are shown in Figure.~\ref{fig:detect_result}.
After that, we adopt random color jitter, random grayscale, random gaussian blur and random horizontal flip to perform data augmentation for each image. In our experiment, the baseline ResNet-50 model achieves 2.16\% improvement in top-1 accuracy by using the above augmentation. One possible reason is that the scale of our target dataset is small and the visual similarity across training samples is high.

For model training, we adopt the ResNet-50~\cite{He2016resnet} model as our backbone in this paper, with the first three stages as the domain-private encoders and the fourth stage as the shared encoder. We append a projection head and a classification head on top of the shared encoder (\ie the last stage of ResNet-50) to perform contrastive learning and classification, respectively. Both the projection head and the classification head are implemented as two fully-connected layers with ReLU activation. The input to the \emph{PSA} module is the middle activated output from the classification head. The encoders are initialized with the ImageNet self-supervised model Dino~\cite{Caron2021EmergingPI} and the domain-private encoders are trained in an end-to-end fashion without pre-training on domain data separately.

We train our model using the Adam optimizer with weight decay set to $10^{-3}$.
We adopt a mini-batch size of $|S^t|=|S^a|=128$ and an initial learning rate of $10^{-4}$. As the auxiliary domain can be much larger than the target domain, we find it to be beneficial by sampling balancedly from the two domains with a ratio of $1:1$. For training stability, we first train our model without the prototypical semantic alignment loss for 5 epochs as a warm-up, then continue training by empirically set the balancing coefficients $\alpha, \beta, \gamma$ in the objective function to $0.1, 0.01, 0.1$, respectively. 
We conduct an ablation study to evaluate the impact of the thresholds $\sigma_{align}$ and $\sigma_{clf}$ for cross-domain knowledge transfer, based on which we set $\sigma_{align}=0.4$ and $\sigma_{clf}=0.9$ in the rest of the experiments.

\begin{table}[t!]
\caption{Ablation studies for loss efficients $\alpha, \beta, \gamma$. }
\begin{tabular}{ccc|cc}
\Xhline{1.3pt}
$\alpha$ & $\beta$   & $\gamma$ & Top-1 & Top-5 \\
\hline
\textbf{0.1}    & \textbf{0.01}  & \textbf{0.1}   & \textbf{87.96} & \textbf{98.69} \\
\hline
0.05   & -    & -      & 86.43 & 98.63 \\
0.2    & -    & -      & 87.75 & 98.80 \\
0.3    & -    & -      & 87.77 & 98.85 \\
0.4    & -    & -      & 87.25 & 98.68 \\
0.5    & -    & -      & 86.63 & 98.69 \\
\hline
-      & 0.001 & -  & 86.89 & 98.65 \\
-      & 0.05  & -  & 87.32 & 98.77 \\
-      & 0.1   & -  & 86.87 & 98.56 \\
-      & 0.2   & -  & 86.96 & 98.44 \\
-      & 0.5   & -  & 87.00 & 98.52 \\
\hline
-    & -      & 0.01       & 86.99 & 98.70 \\
-    & -      & 0.05      & 87.19 & 98.16 \\
-    & -      & 0.2        & 87.08 & 98.57 \\
-    & -      & 0.3     & 86.74 & 98.60 \\
-    & -      & 0.4      & 86.82 & 98.62 \\
\Xhline{1.3pt}
\end{tabular}
\label{table:abla_3loss}
\end{table}

%\textbf{Training.}

\subsection{Ablation Studies for Loss Coefficients}
Here we conduct three ablation studies for $\alpha, \beta, \gamma$ for $\mathcal{L}_{ada}$, $\mathcal{L}_{psa}$, $\mathcal{L}_{inter\_clf}$, respectively, on the Visda-2017 dataset due to its more stable results from a large testset.
As shown in Table~\ref{table:abla_3loss}, we compare the top-1 and top-5 accuracy of one candidate by holding the other two fixed as the best value we found during the experiment, \emph{i.e.}, $\alpha=0.1$, $\beta=0.01$ and $\gamma=0.1$.
We can see that even though our method achieved better top-1 and top-5 accuracies regardless of the coefficients setting compared with the existing solutions as shown in our main text, a proper one still serves as an important factor for further improvement.

\end{document}